\documentclass[lettersize,journal]{IEEEtran}
\usepackage{amsmath,amsfonts}
\usepackage{algorithmic}
\usepackage{algorithm}
\usepackage{array}
\usepackage{textcomp}
\usepackage{stfloats}
\usepackage{url}
\usepackage{verbatim}
\usepackage{cite}
\usepackage{graphicx}
\usepackage{subfigure}
\graphicspath{{pdfs/},{images/} }

\usepackage{multirow}

\usepackage{booktabs}

\usepackage{cleveref}

\usepackage{makecell}

\usepackage{amssymb}

\usepackage{threeparttable}

\newcommand{\figref}[1]{Fig.~\ref{#1}}
\newcommand{\tabref}[1]{Table~\ref{#1}}
\newcommand{\secref}[1]{Section~\ref{#1}}

\newcommand{\algref}[1]{Algorithm~\ref{#1}}

\makeatletter
\renewcommand{\maketag@@@}[1]{\hbox{\m@th\normalsize\normalfont#1}}%
\makeatother

\begin{document}

\title{SP-VIO: Robust and Efficient Filter-Based Visual Inertial Odometry with State Transformation Model and Pose-Only Visual Description}

\author{Xueyu Du$^{\dag}$, Lilian Zhang$^{\dag}$, Chengjun Ji, Xinchan Luo, Huaiyi Zhang, Maosong Wang, Wenqi Wu, and Jun Mao$^*$
\thanks{Authors are with the College of Intelligent Science and Technology, National University
	of Defense Technology, Changsha, 410073, China.}
\thanks{$^{\dag}$ These authors contributed equally to this work.}
\thanks{$^*$ Correspondence to: maojun12@nudt.edu.cn.}
\thanks{This research is funded by the National Natural Science Foundation of China (grant number: 62103430, 62103427, 62073331) and Major Project of Natural Science Foundation of Hunan Province (No. 2021JC0004).}}

\markboth{Journal of \LaTeX\ Class Files,~Vol.~14, No.~8, August~2021}%
{Shell \MakeLowercase{\textit{et al.}}: A Sample Article Using IEEEtran.cls for IEEE Journals}

\maketitle

\begin{abstract}
Due to the advantages of high computational efficiency and small memory requirements, filter-based visual inertial odometry (VIO) has a good application prospect in miniaturized and payload-constrained embedded systems. 
However, the filter-based method has the problem of insufficient accuracy.
To this end, we propose the State transformation and Pose-only VIO (SP-VIO) by rebuilding the state and measurement models, and considering further visual deprived conditions.
In detail, we first proposed the double state transformation extended Kalman filter (DST-EKF) to replace the standard extended Kalman filter (Std-EKF) for improving the system's consistency, and then adopt pose-only (PO) visual description to avoid the linearization error caused by 3D feature estimation.
The comprehensive observability analysis shows that SP-VIO has a more stable unobservable subspace, which can better avoid the inconsistency problem caused by spurious information.
Moreover, we propose an enhanced double state transformation Rauch-Tung-Striebel (DST-RTS) backtracking method to optimize motion trajectories during visual interruption.

Monte-Carlo simulations and real-world experiments show that SP-VIO has better accuracy and efficiency than state-of-the-art (SOTA) VIO algorithms, and has better robustness under visual deprived conditions.

\end{abstract}


\section{Introduction}\label{sec:Introduction}
\IEEEPARstart{V}{isual} inertial odometry (VIO) technology is widely used in various unmanned systems for autonomous navigation. Although different VIO algorithms have been developed in the past decade, the unmanned systems industry still expect algorithms with lower computation costs, higher accuracy, and better robustness.

The mainstream VIO algorithms can be broadly divided into two categories: optimization-based \cite{WOS:000350472800005,WOS:000712319501044,WOS:000442341000003,WOS:000725804900006,WOS:000525362000002} and filter-based methods \cite{4209642,9196524,doi:10.1177/0278364913481251,WOS:000424646100016,WOS:000411059400002,indelman2012real,fan2024schurvins}.
Benchmark experiments \cite{WOS:000446394502008,WOS:000494942307003} show that filter-based VIO has the advantages of high computational efficiency and small memory requirements, and has a good application prospect in payload-constrained embedded systems. 
However, compared with the optimization-based VIO, filter-based methods are more efficient but also have the problem of insufficient accuracy, which limits their application in some scenarios.

Taking MSCKF \cite{4209642,9196524,doi:10.1177/0278364913481251,WOS:000424646100016}, the representative work of filter-based VIO, as an example, this method uses null space projection to simplify the residual model and then performs EKF update to estimate the corresponding pose.  
Because the 3D features used in the visual residual model are obtained only through a 3D reconstruction process and are not jointly optimized with the system's pose, it is easy to cause the accumulation of linearization errors and lead to the decline of navigation accuracy.
In addition, EKF is prone to filtering inconsistency problems, that is, misestimating the observability of the system leads to performance degradation, which is also an important factor affecting the accuracy of filter-based VIO \cite{doi:10.1177/0278364913481251}.

In summary, it can be found that the main reason for the accuracy damage of filter-based VIO is the accumulation of linearization errors \cite{fan2024schurvins}. Therefore, if we can solve the problem that filter-based methods are susceptible to linearization errors, it will help to develop a VIO framework that is both highly precise and efficient.

As mentioned above, an important problem of MSCKF-based VIO is that it is not entirely decoupled from 3D features, which are still required to construct visual residuals.
The newly proposed PO theory indicates that reprojection error can be obtained solely through camera pose, equivalent to traditional multi-view geometry description \cite{WOS:000458768000003,WOS:000899419900005}. 
Inspired by PO theory, we derive a visual residual model based only on pixel coordinates and relative pose, which is completely decoupled from 3D features to avoid being affected by inaccurate 3D reconstruction processes. Meanwhile, the new residual model does not need the simplification of left null space, and can be used for filter update directly.

On the other hand, the current strategies to solve the problem of filtering inconsistency can be broadly classified into two categories. One starts from the perspective of maintaining the system observability, such as the first estimate Jacobian EKF (FEJ-EKF) \cite{FEJ,doi:10.1177/0278364913481251,FEJ2} and the observability constrained EKF (OC-EKF) \cite{doi:10.1177/0278364909353640,6605544}.
Since different scenes have diverse effects on the system observability, such as static and dynamic scenes, thus this method is more demanding in practical applications \cite{wang2018state}. 
The other one starts with the model definition, such as the invariant Kalman filter (IEKF) \cite{9689974,7523335,hartley2020contact,9669116,10179117}, which uses the matrix Lie group to define the state space more strictly. The advantage of IEKF is its universality in different scenarios, but the derivation and implementation are too complex \cite{wang2018state}.
ST-EKF is a special case of matrix Lie group application, which obtains accuracy improvement by state transformation of velocity error and is easier to implement than IEKF \cite{wang2018state,wang2019consistent,wang2021schmidt}.
However, ST-EKF only has a more rigorous definition of the velocity error state, and there are still theoretical imperfections. 
Based on ST-EKF, we propose the double state transformation EKF (DST-EKF), which redefines the system's position error state and further improves the accuracy and consistency.

The above works are concerned with solving the accumulation of linearization errors caused by inaccurate system modeling. In addition, visual measurement information is also an important source of linearization errors.
Due to the camera's sensitivity to the environment, it is easy to be disturbed in the actual motion, resulting in the visual tracking failure, which makes the system input discontinuous measurement information and affects the subsequent state estimation accuracy.
The current common solution is the visual relocalization \cite{10475537}, but this method relies on matching with the existing localization scenes and does not work when the motion trajectory is not repeated, which limits the robustness of VIO under non-closed-loop motion trajectories.
Rauch-Tung-Striebel (RTS) \cite{rts} is an efficient optimal backtracking smoothing method, which can optimize the cumulative error based on the historical state information, and is not limited by whether a closed-loop trajectory is formed.
Therefore, we propose the DST-RTS backtracking smoothing method combining DST-EKF and RTS smoother to correct motion trajectories during visual interruption.
This will improve the robustness of VIO system under visual deprived conditions.

\begin{figure*}[htbp]
	\centering
	\subfigure[] { \centering \includegraphics [width=2.3in] {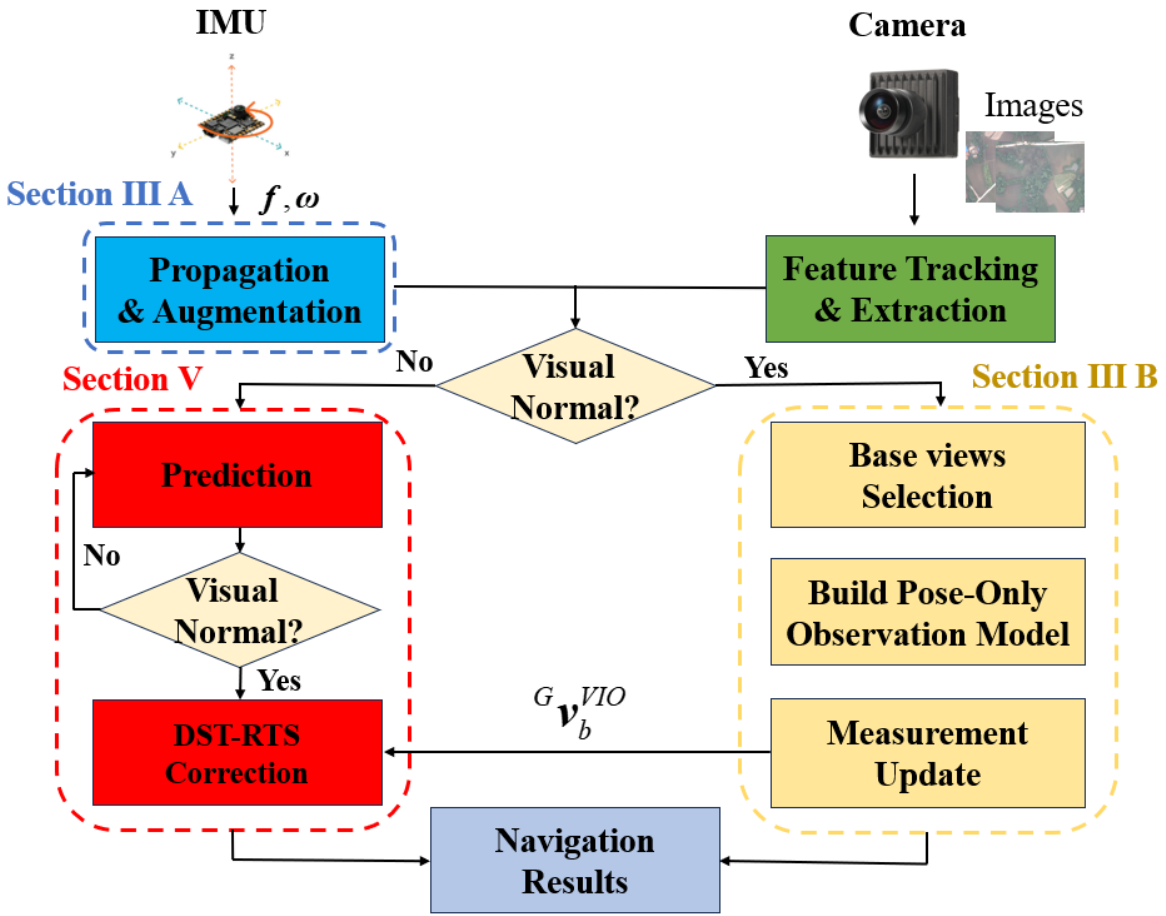}\label{fig:Overview1}} 
	\subfigure[] { \centering \includegraphics [width=2.3in] {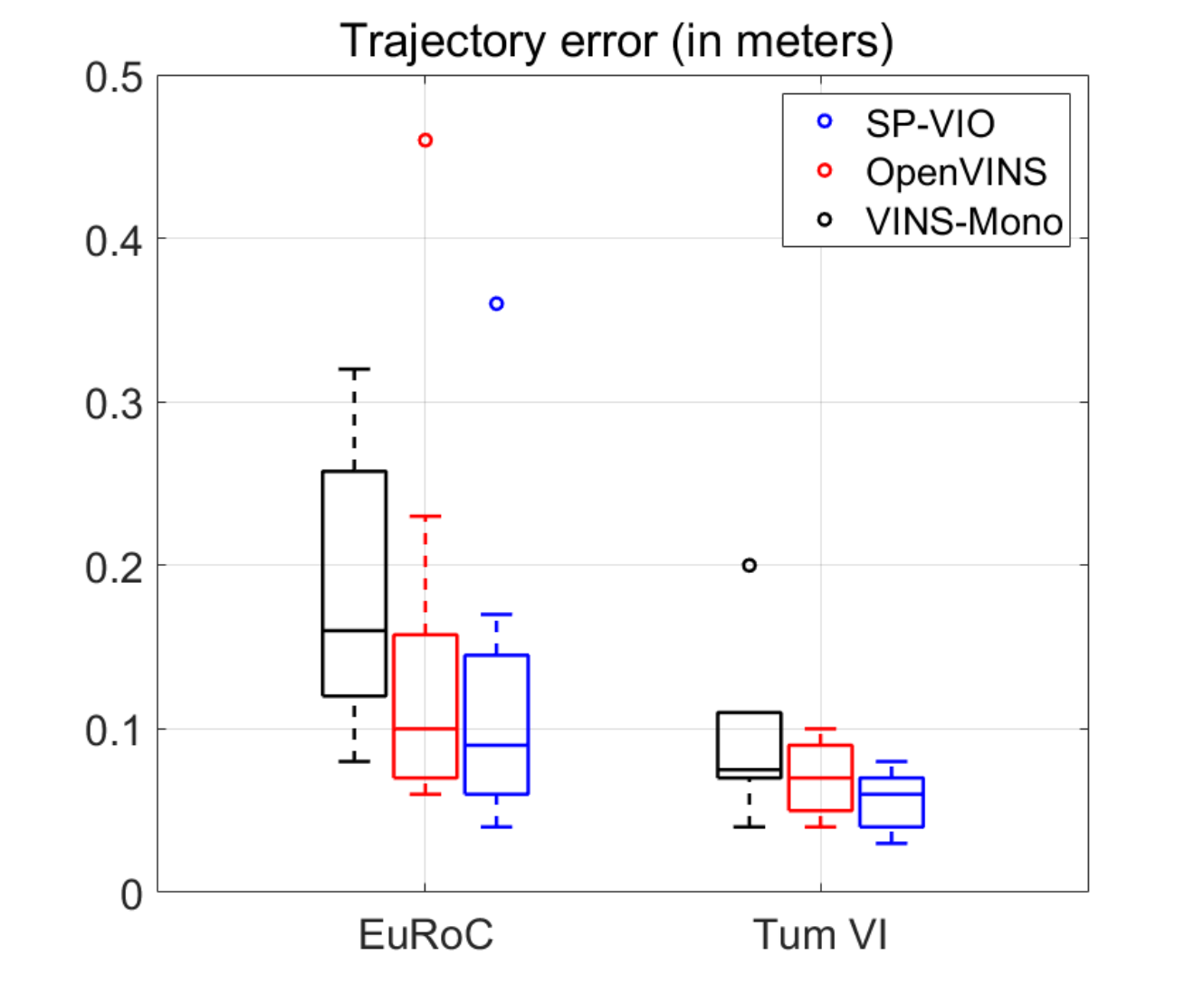}\label{fig:Overview2}}
	\subfigure[] { \centering \includegraphics [width=2.3in] {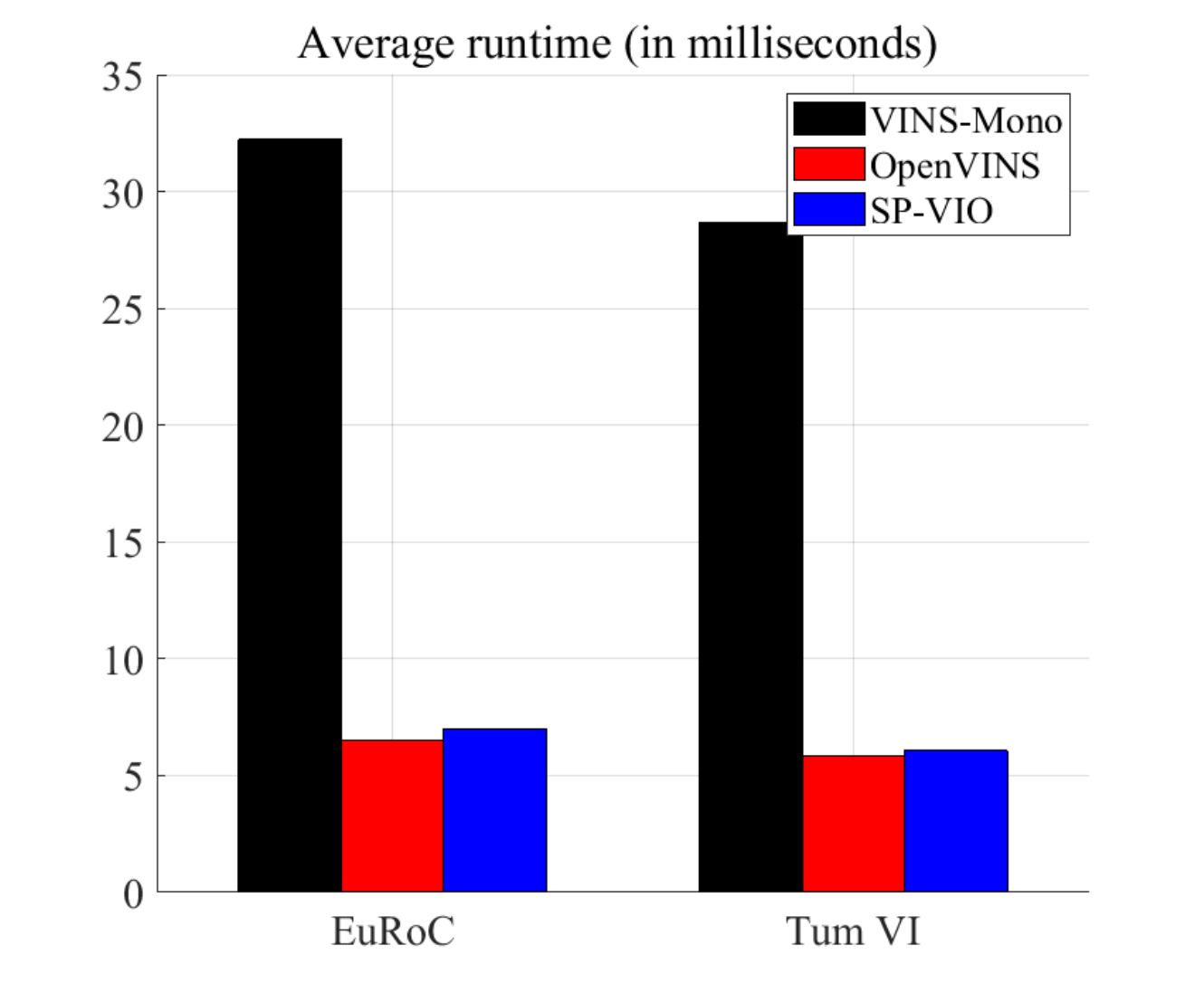}\label{fig:Overview3}}
	\caption{
		This figure illustrating the full pipeline of the proposed SP-VIO, and part of the performance test results of this algorithm on popular public datasets, including trajectory errors and time costs.
		(a) Framework of SP-VIO.
		(b) Trajectory error on EuRoC \cite{WOS:000382981300001} and Tum VI \cite{8593419}.
		(c) Average runtime on EuRoC and Tum VI.
	}
	\label{fig:Overview}
\end{figure*}

In conclusion, we propose the state transformation pose-only VIO (SP-VIO) algorithm, as shown in \figref{fig:Overview}, which considers both high accuracy and efficiency, and has certain robustness under visual deprived conditions.
To be specific, the contributions of this paper are shown as follows:

\begin{itemize}
        \item We propose a novel VIO system named SP-VIO. Specifically, we first propose the DST-EKF to replace the Std-EKF for improving the system's consistency by more strict velocity and position error definitions. Meanwhile, we adopt PO visual description to decouple the measurement model from 3D features to avoid the linearization error caused by inaccurate feature triangulation.	
        \item We, for the first time, analyze the PO visual description based VIO system's observability, and demonstrate that VIO system combining DST-EKF and PO-based measurement model have more stable unobservable subspace, thus better avoiding inconsistency problems due to linearization errors.
	\item We propose a DST-RTS backtracking smoothing strategy that does not rely on loop closure. This strategy uses the velocity information provided by the VIO system after the visual observation is recovered to correct the motion trajectory by RTS backtracking, which can reduce the cumulative error caused by visual interruption.
	\item We conduct comprehensive performance evaluations through Monte-Carlo simulations and real-world experiments, including popular public datasets (EuRoC\cite{WOS:000382981300001}, Tum VI\cite{8593419}, Kitti Odometry\cite{WOS:000309166203066}) and personal datasets. 
	Experimental results show that SP-VIO has both high precision and robustness, with comprehensive performance better than SOTA VIO algorithms.
\end{itemize}

\section{Related Work}\label{sec:Related Work}
\subsection{Visual Inertial Odometry Technology}\label{sec:Visual Inertial Odometry Technology}
Although VIO and Visual-Inertial Simultaneous Localization And Mapping (VI-SLAM) both involve the fusion application of visual information and inertial data, compared with VI-SLAM, VIO focuses more on its pose estimation rather than long-term mapping or relocalization performance.
According to the state estimation method, the mainstream VIO algorithms can be roughly divided into filter-based and optimization-based \cite{WOS:000494942307003}. 

The early filter-based methods have been mainly developed on EKF-SLAM \cite{book}, which performs joint state estimation of imu pose and feature positions; the disadvantage of this method is prone to the problem of feature dimension explosion \cite{doi:10.1177/0278364913481251}.
As one of the most classic filter-based methods, MSCKF \cite{4209642} has proposed to remove the feature information from the filter state space and constructs the multi-view constraints in the observation model, which greatly reduces the computational complexity compared to EKF-SLAM \cite{doi:10.1177/0278364913481251}.

The original MSCKF has been extended and improved in different ways, such as observability constraints \cite{FEJ,doi:10.1177/0278364913481251}, square-root form \cite{WOS:000570976400008, 7989022}, and the application of the Lie group theory. \cite{doi:10.1146/annurev-control-060117-105010,10179117}. Representative works of filter-based VIO also include ROVIO \cite{WOS:000411059400002}, trifocal-EKF \cite{indelman2012real}, and SchurVINS \cite{fan2024schurvins}.

The optimization-based methods use batch graph optimization \cite{grisetti2011g2o} or bundle adjustment techniques \cite{ceres-solver} to obtain the optimal state estimation.
To achieve a constant processing time, the optimization-based VIO typically only considers the bounded sliding window of recent states as active optimization variables, while marginalizing past states and measurements \cite{WOS:000280869700005}.
In \cite{6630883} and \cite{9873988}, efficient Hessian construction and Schur complement calculation are employed to improve cache efficiency and avoid redundant matrix representation.
For inertial constraints, the proposal of pre-integration theory \cite{WOS:000300188300006} has allowed high-rate IMU measurements to be effectively integrated into the optimization process.

Leutenegger has proposed a keyframe-based VIO algorithm \cite{WOS:000350472800005}, which maintains a finite size optimization window by marginalizing old keyframes.
VINS-Mono proposed by Qin \cite{WOS:000442341000003} uses sliding windows for nonlinear optimization, which can achieve online spatial calibration and loop closures.
The current mainstream optimization-based VIO algorithms also include Kimera \cite{WOS:000712319501044}, BASALT \cite{WOS:000525362000002}, VI-DSO \cite{WOS:000446394502009}, SVO2 \cite{WOS:000399348900001}, and ORB-SLAM3\cite{WOS:000725804900006}.

Benchmark comparisons about a series of state-of-the-art VIO systems have shown that \cite{WOS:000446394502008}, optimization-based VIO has better localization accuracy but consumes more computing resources; while filter-based VIO is more suitable for miniaturized systems with limited resources due to its high efficiency, but its localization performance is more susceptible to linearization errors.
Therefore, an important development direction of VIO technology will be constructing a framework that combines high accuracy and efficiency.

\begin{figure*}[htbp]
	\centering
	\subfigure[] { \centering \includegraphics [width=2.3in] {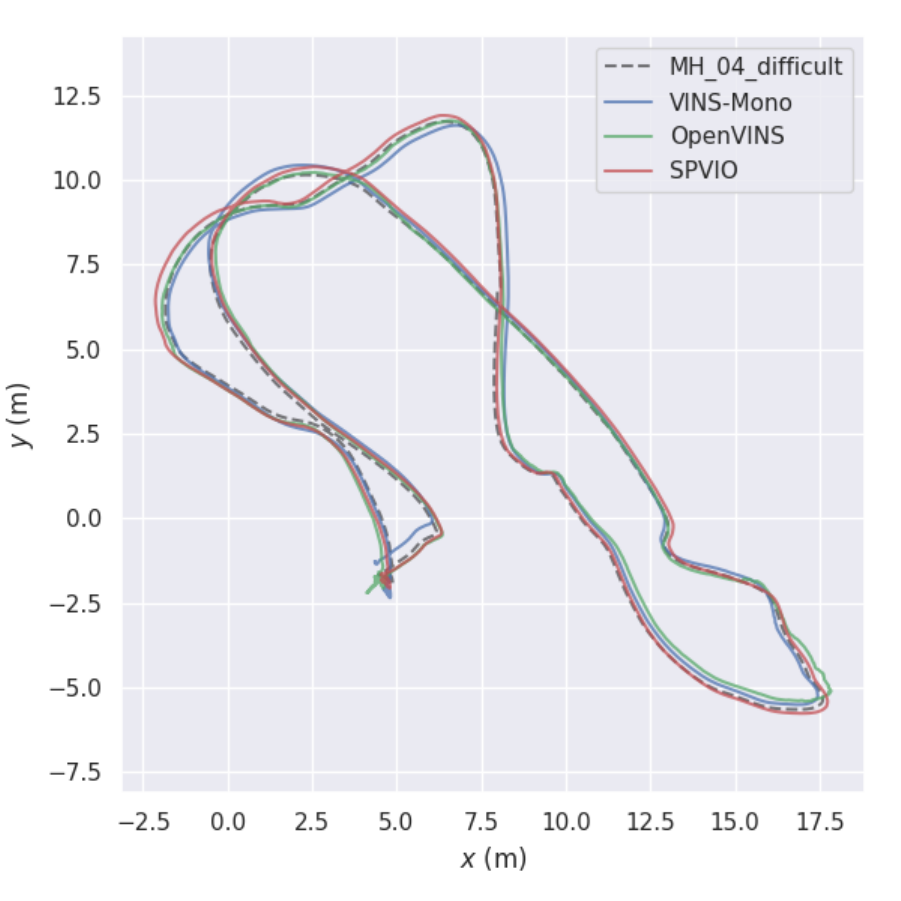}\label{fig:eurocmh041}} 
	\subfigure[] { \centering \includegraphics [width=2.3in] {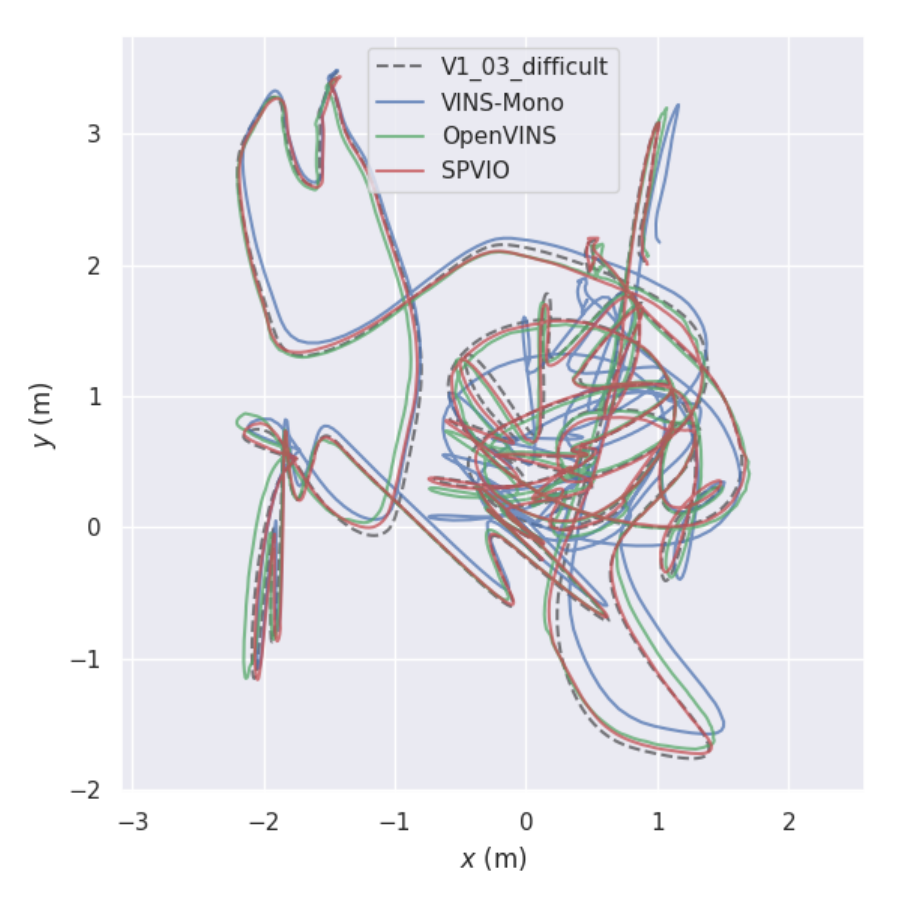}\label{fig:tum051}}
	\subfigure[] { \centering \includegraphics [width=2.3in] {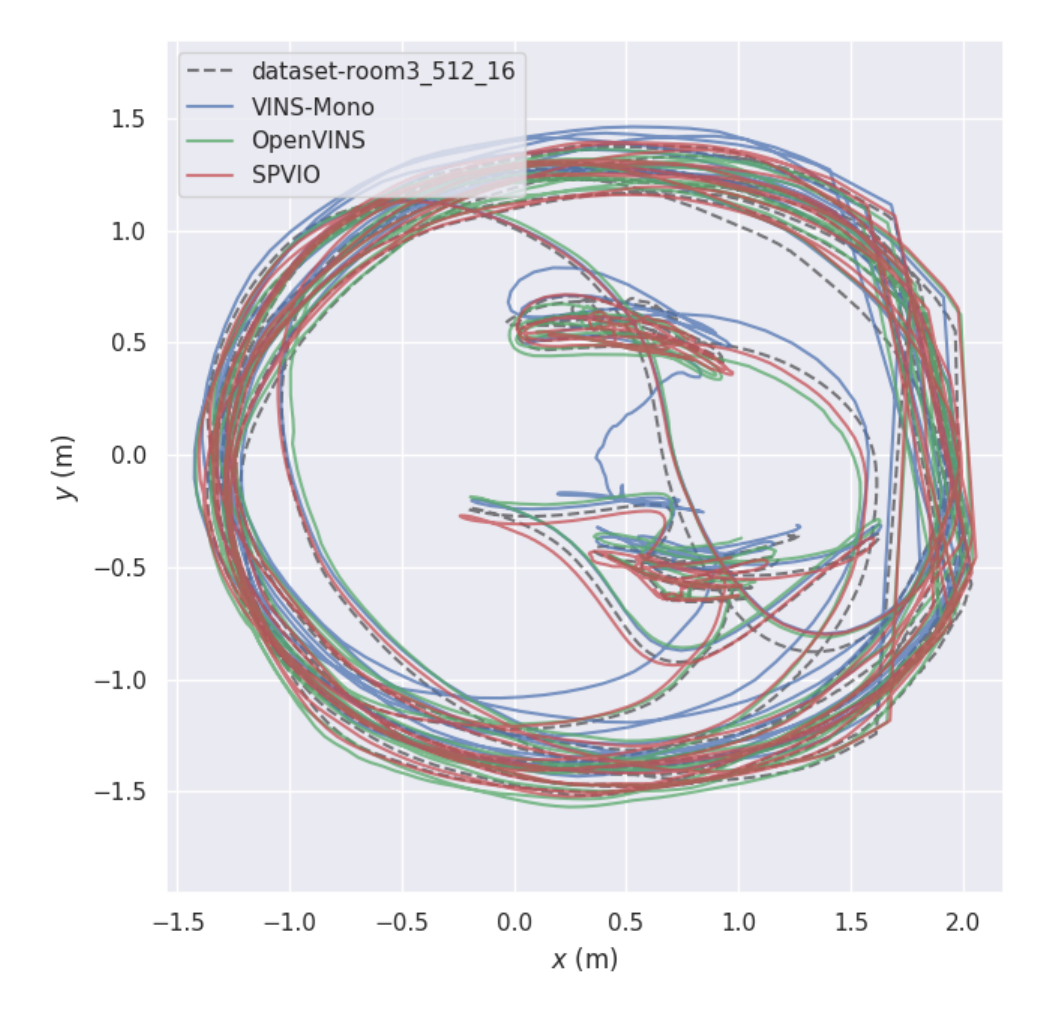}\label{fig:kitti071}}
	\caption{
		Partial result trajectories in public datasets, compared with OpenVINS and VINS-Mono.
		(a) EuRoC MH\_04\_difficult.
		(b) EuRoC V1\_03\_difficult.
		(c) Tum VI Room3\_512\_16.
	}
	\label{fig:Partial result trajectories in public datasets2}
\end{figure*}

\subsection{Visual Feature}\label{sec:Visual feature representation}
Visual feature is an important part of VIO technology, which is related to the accuracy and computational efficiency of the systems.
According to geometric properties, visual features can be divided into point, line and plane features \cite{8799000}, and point features are commonly used in the mainstream VINS algorithm \cite{WOS:000442341000003,WOS:000725804900006,9196524}. 
However, in the weak texture environment, line and plane features are more robust and thus are used to maintain the localization accuracy \cite{10160620,9341278,8967905,8731724}.
Nevertheless, point features are still a common and effective choice for VINS algorithms in terms of efficiency and generality.

The VIO algorithms adopt the visual feature's 3D position to build visual residual model, which can be used to optimize the estimated state.
The 3D feature can be obtained through a 3D reconstruction process that minimizes the reprojection error, mainly achieved by inverse depth parameterization\cite{4637878} and Bundle Adjustment (BA) \cite{10.5555/1888028.1888032}.
To achieve higher accuracy, 3D feature is often used as a long-term estimate state in VIO, which increases the computational overhead.
For higher efficiency, MSCKF \cite{4209642} no longer puts 3D features into the state space for joint optimization with the IMU state, which results in its accuracy being limited by the 3D reconstruction process.

The newly proposed PO theory indicates that traditional multi-view geometric description can be equivalently represented by the pose-only constraints \cite{WOS:000458768000003,WOS:000899419900005}.
In the new description, the visual residual is decoupled from the 3D features and only related to the camera pose and 2D features.
At present, this theory has been applied in optimization-based VO \cite{WOS:000899419900005} and VIO \cite{10438889}, while the localization accuracy and efficiency have been greatly improved.
Therefore, PO theory can also be used to construct the MSCKF-based VIO to decouple the observation model from the 3D features, while maintaining efficiency and also avoiding the influence of an inaccurate 3D reconstruction process.

\begin{figure*}[htbp]
	\centering
	\subfigure[VINS-Mono] { \centering \includegraphics [width=2.3in] {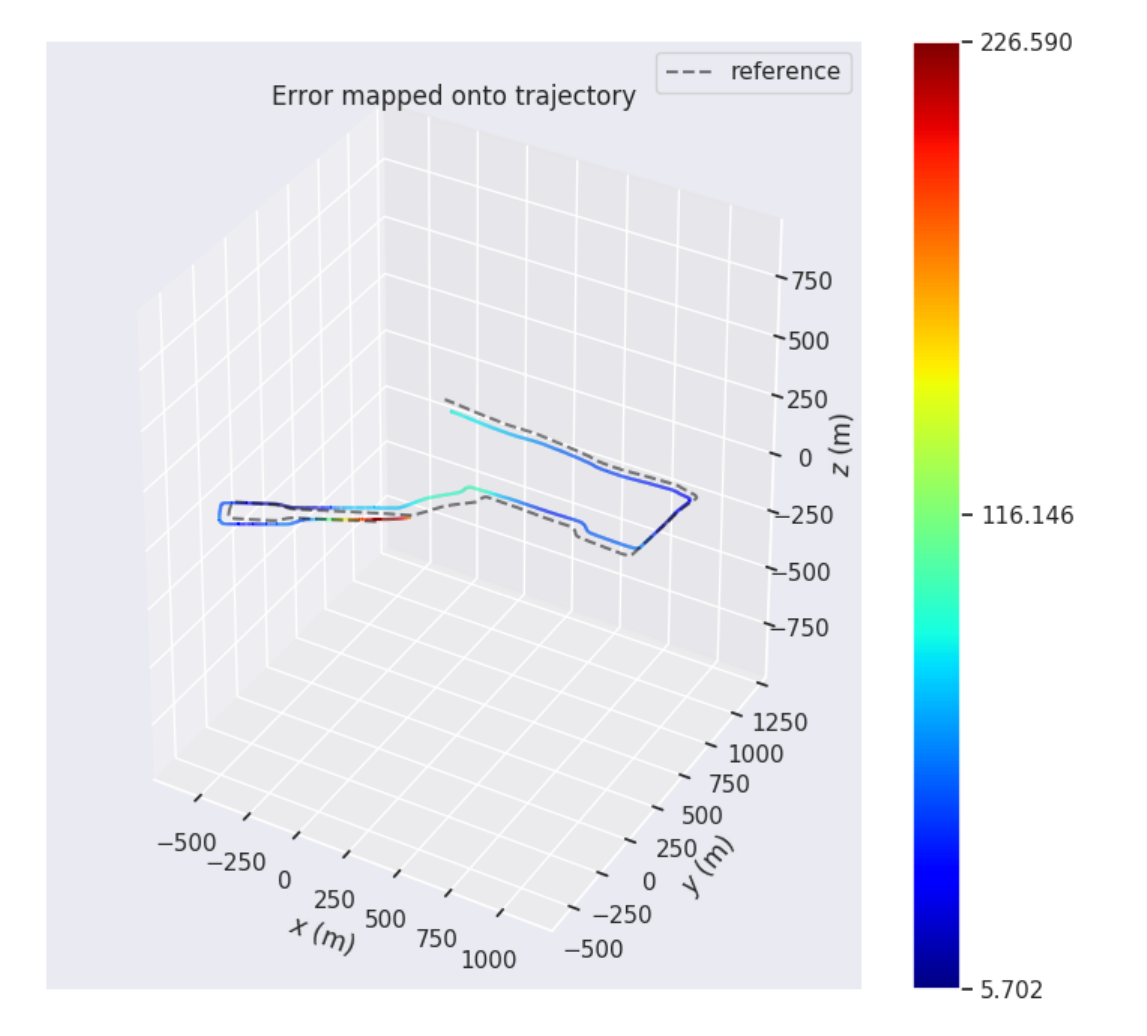}\label{fig:vinsmonocar}} 
	\subfigure[OpenVINS] { \centering \includegraphics [width=2.3in] {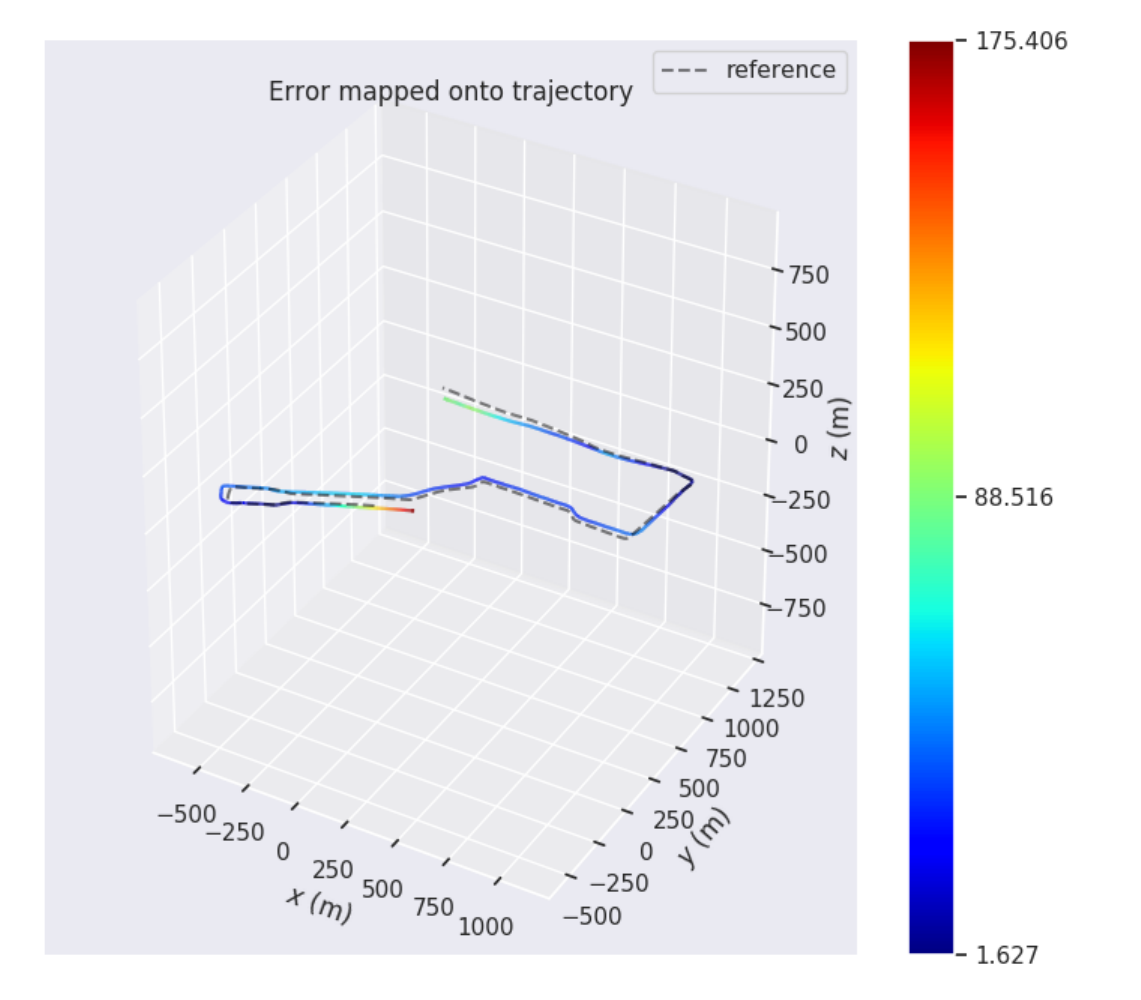}\label{fig:openvinscar}}
	\subfigure[SP-VIO] { \centering \includegraphics [width=2.3in] {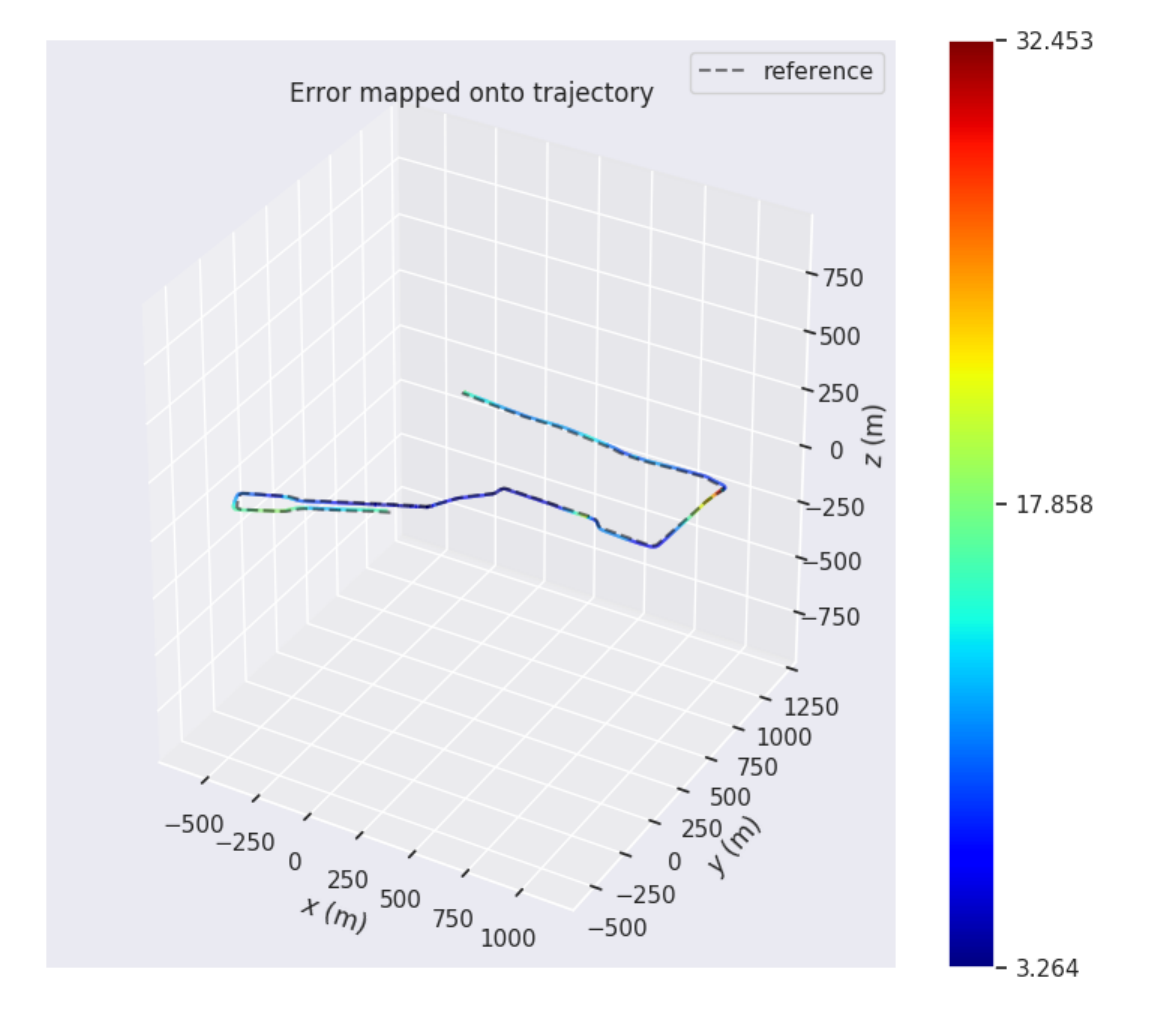}\label{fig:SPVIOcar}}
	\caption{
		Comparison of the estimated trajectory of three algorithms with the ground truth in nudt\_car, the total distance of the trajectory is 4483.54m, and the percentage of localization error on VINS-Mono, OpenVINS, and SP-VIO are 1.62\%, 1.11\%, and 0.29\%, repectively.
	}
	\label{fig:comparison in nudt-vi car}
\end{figure*}

\subsection{Filtering Inconsistency Problem}\label{sec:Inconsistency problem of Filter-based VIO}
In state estimation, if the estimation error has zero mean and the covariance is less than or equal to the calculated covariance, then the filtered estimate is consistent. Consistency is one of the main criteria to evaluate the performance of an estimator \cite{book2}, and is closely related to observability.
Huang \cite{doi:10.1177/0278364909353640} demonstrated that the inconsistency of standard EKF SLAM is associated with a mismatch between the observability of the linearized error-state system and the observability of the true system.

The current strategies to solve the inconsistency problem of filter-based VIO can be broadly classified into two categories.
The first class of methods is to maintain the system's observability consistent with the reality.
Geneva \cite{geneva2019efficient} and Tong \cite{ke2019rise} combine the Schmidt Kalman filter (SKF) with the visual-inertial navigation system (VINS), which sets the Kalman filter gain corresponding to the unobservable state to zero, that is, the unobservable state is not corrected to ensure the consistency of the system.
Hesch \cite{6605544} developed an observability-constrained EKF (OC-EKF) for VIO that directly enforces the unobservable directions of the system in the update step of the EKF and showed that this significantly reduces filter inconsistency.
Huang proposed the first-estimates Jacobian extended Kalman filter (FEJ-EKF)\cite{FEJ}, which reduces the damage of the observability problem to the consistency of EKF. Li and Mourikis \cite{doi:10.1177/0278364913481251} applied the FEJ concept to an MSCKF design and showed that this improved consistency over a standard MSCKF.
To solve the problem that FEJ is sensitive to poor initial estimates, chen proposed the FEJ2 method which is robust to large initialization errors\cite{FEJ2}.

The second type of method exploits the Lie group symmetry to redefine the system error state, which improves the system's consistency, such as the invariant EKF (IEKF)\cite{7523335} and equivariant filter(EQF)\cite{9840886}.
Barrau and Bonnabel proposed a new Lie group $S{E}_{n}(m)$ and showed that this is a symmetry suitable for the classical SLAM problem \cite{barrau2015ekf}. 
They further proposed IEKF \cite{7523335} based on this method, which overcame the consistency problem that plagued EKF-based SLAM \cite{bailey2006consistency}.
Zhang \cite{7812660} performed an observability analysis of IEKF-SLAM and demonstrated the advancement of the method through simulations.
\cite{8205965,8593627,8417453,8299463,hartley2020contact,9669116} applied IEKF and its variants to MSCKF-based VIO and achieved consistency improvement.
For SLAM systems with equivariant visual measurements \cite{van2021constructive}, Goor proposed the EQF \cite{9840886}, which also provides a framework to reduce linearization errors by using equivariance of system output functions. 
EQF has been further applied in the VIO system, which shows good positioning accuracy and robustness \cite{10179117}.

The disadvantage of the first kind of method is that it must be based on a deep understanding of the system observability, and cannot solve the inconsistency problem caused by the system modeling from the root. The second method does not need to know the incomplete observable scene of the system in advance, so it has better engineering practicability than the first method, but the disadvantage is that the filter design process is too complex \cite{wang2018state}. 
Based on IEKF\cite{7523335}, Wang proposed the state transformation Kalman filter (ST-EKF), which improved the consistency and accuracy by redefining the velocity error \cite{wang2018state,wang2019consistent,wang2021schmidt}. Compared with IEKF, the filter design of ST-EKF is simpler and easier to implement.
However, since the position error state of ST-EKF is still defined traditionally, there is still space for improvement in terms of the system model.
Therefore, continuing the state transformation for the position error state based on the ST-EKF is helpful to further improve the consistency of the VIO system, while also retaining the advantage of easy implementation of ST-EKF.

\subsection{Correction Strategy under Visual Deprived Conditions}\label{sec:Visual Relocation Strategy under Discontinuous Observation}
In the case of visual short-term tracking failure (less than 5s), VINS usually uses the pose predicted by the IMU to match the map keyframe in a larger image window and performs pose correction if the match is successful \cite{WOS:000442341000003,WOS:000725804900006}.
When the visual interruption lasts for a long time, a new map is generated after the observation recovery and is fused with the map built before the tracking failure through visual relocalization to compensate for the localization errors \cite{WOS:000442341000003,WOS:000725804900006}.

The visual relocalization problem \cite{10475537} is a key component in visual navigation applications, designed to optimally estimate pose when a device revisits an area in a map.
Traditional visual relocalization technology can be divided into image-to-image and image-to-map.
Image-to-image is to obtain pose through the matching relationship between the current frame image and the map key frame image. At present, the method based on the Bag of Words (BoW) model \cite{WOS:000186833000193} is widely used, which has been successfully applied to PTAM\cite{4538852} and ORB-SLAM\cite{4538852}.
Image-to-map is the construction of a point cloud map through 3D reconstruction, which is then correlated with local features in the observed image. Hierarchical localization (HLoc) \cite{8953492} is one of the most widely used methods at present. It uses neural network for retrieval and local 2D-3D matching, which can achieve high precision and high robustness localization in large scenes.

The above methods depend on matching current and historical scenes. If no similar scene is detected, the localization error accumulated during the visual interruption cannot be corrected, which limits the application of relocalization method. 
For example, in the large scene of the UAV flight task, it is difficult to go through repeated scenes, which cannot be applied to the traditional relocalization method.
Therefore, for non-closed loop motion under visual deprivation, a correction strategy independent of scene matching is needed.

\section{Basic SP-VIO Framework}\label{sec:Barebones SP-VIO}
In this paper, the basic SP-VIO is developed based on OpenVINS\cite{9196524}, where the DST-EKF is used to replace the Std-EKF, and the measurement model based on Pose-only visual description is utilized to replace the original multi-view geometry based measurement model. The framework of SP-VIO algorithm is shown in \figref{fig:Overview1}.

\subsection{System Model Based on DST-EKF}\label{sec:System model}
\subsubsection{State Definition}\label{sec:State Definition}
The state vector of our visual-inertial system consists of the current inertial navigation state, and a set of historical IMU pose clones.
\begin{align}
	\boldsymbol{x}_{k}&=\begin{bmatrix} \boldsymbol{x}_{b}^{T} & \boldsymbol{x}_{c}^{T} \end{bmatrix} ^{T} \label{eqn:x state}\\
	\boldsymbol{x}_{b}&=\begin{bmatrix} _{{b}_{k}}^{G}\boldsymbol{q}^{T} & ^{G}\boldsymbol{v}_{{b}_{k}}^{T} & ^{G}\boldsymbol{p}_{{b}_{k}}^{T} & \boldsymbol{b}_{{g}_{k}}^{T} & \boldsymbol{b}_{{a}_{k}}^{T} \end{bmatrix} ^{T} \label{eqn:xb state}\\
	\boldsymbol{x}_{c}&=\begin{bmatrix} _{{b}_{k-N}}^{G}\boldsymbol{q}^{T} & ^{G}\boldsymbol{p}_{{b}_{k-N}}^{T} & {...} & _{{b}_{k-1}}^{G}\boldsymbol{q}^{T} & ^{G}\boldsymbol{p}_{{b}_{k-1}}^{T} \end{bmatrix} ^{T} \label{eqn:xc state}
\end{align}
where $_{{b}_{k}}^{G}\boldsymbol{q}$ represents the rotation unit quaternion from the body frame $\{b\}$ to the global frame of refence $\{G\}$ at time $k$, $\boldsymbol{b}_{{g}_{k}}$ and $\boldsymbol{b}_{{a}_{k}}$ are the gyroscope and accelerometer biases, and $^{G}\boldsymbol{v}_{{b}_{k}}$ and $^{G}\boldsymbol{p}_{{b}_{k}}$ are the velocity and position of the body frame expressed in the global frame.
From Eq. \eqref{eqn:x state} to Eq. \eqref{eqn:xc state}, the corresponding error-state of $\boldsymbol{x}_{k}$ can be defined as:
\begin{align}
	\delta \boldsymbol{x}_{k}&=\begin{bmatrix} \delta \boldsymbol{x}_{b}^{T} & \delta \boldsymbol{x}_{c}^{T} \end{bmatrix} ^{T} \label{eqn:delta x state}\\
	\delta \boldsymbol{x}_{b}&=\begin{bmatrix} _{{b}_{k}}^{G}\boldsymbol{\phi}^{T} & \delta ^{G}\boldsymbol{v}_{{b}_{k}}^{T} & \delta ^{G}\boldsymbol{p}_{{b}_{k}}^{T} & \delta \boldsymbol{b}_{{g}_{k}}^{T} & \delta \boldsymbol{b}_{{a}_{k}}^{T} \end{bmatrix} ^{T} \label{eqn:delta xb state}\\
	\delta \boldsymbol{x}_{c}&=\begin{bmatrix} _{{b}_{k-N}}^{G}\boldsymbol{\phi}^{T} & \delta ^{G}\boldsymbol{p}_{{b}_{k-N}}^{T} & {...} & _{{b}_{k-1}}^{G}\boldsymbol{\phi}^{T} & \delta ^{G}\boldsymbol{p}_{{b}_{k-1}}^{T} \end{bmatrix} ^{T} \label{eqn:delta xc state}
\end{align}
where $_{{b}_{k}}^{G}\boldsymbol{\phi}$ represents the error state of $_{{b}_{k}}^{G}\boldsymbol{q}$. Except for quaternion, other states can be used with the error definition (e.g. $\boldsymbol{x} = \boldsymbol{\widetilde{x}} + \delta \boldsymbol{x}$). The hamiton quaternion is used in this paper, and its extended additive error can be defined as:
\begin{align}\label{eqn:quaternions}
	\boldsymbol{q}=\delta \boldsymbol{q} \otimes \boldsymbol{\widetilde{q}}\approx\begin{bmatrix}
		1 \\ \frac{1}{2} \boldsymbol {\phi} 
	\end{bmatrix} \otimes  \boldsymbol{\widetilde{q}}
\end{align}
where $\otimes$ represents quaternion multiplication. Thus the extended additive error of rotation matrix is defined as:
\begin{align}\label{eqn:quaternions to R}
	\boldsymbol{R}(_{{b}_{k}}^{G}\boldsymbol{q}) &= _{{b}_{k}}^{G}\boldsymbol{R} \\  _{{b}_{k}}^{G}\boldsymbol{\widetilde{R}} &= (\boldsymbol{I}-[_{{b}_{k}}^{G}\boldsymbol{\phi} \times])_{{b}_{k}}^{G}\boldsymbol{R}
\end{align}

\subsubsection{Propagation and Augmentation}\label{sec:Propagation and Augmentation}
The traditional velocity error is defined as $\delta ^{G}\boldsymbol{v}_{{b}_{k}} = {^{G}\boldsymbol{v}_{{b}_{k}}}      - {^{G}\boldsymbol{\widetilde{v}}_{{b}_{k}}} $, only the difference numerical values is considered, and the difference between the coordinate systems is not considered. The velocity error redefined by the state transformation method is expressed as follows\cite{wang2018state}:
\begin{align}\label{eqn:delta v new}
	\delta _{ST}^{G}\boldsymbol{v}_{{b}_{k}}&= {^{G}\boldsymbol{v}_{{b}_{k}}} - {_{{b}_{k}}^{G}\boldsymbol{R}}{_{G}^{{b}_{k}}\boldsymbol{\widetilde{R}}}{^{G}\boldsymbol{\widetilde{v}}_{{b}_{k}}} \\
	&\approx - \delta ^{G}\boldsymbol{v}_{{b}_{k}} + [{^{G}\boldsymbol{\widetilde{v}}_{{b}_{k}}}\times]_{{b}_{k}}^{G}\boldsymbol{\phi}    \nonumber
\end{align}

Similarly, the traditional position error definition $\delta ^{G}\boldsymbol{p}_{{b}_{k}} =       {^{G}\boldsymbol{\widetilde{p}}_{{b}_{k}}} - {^{G}\boldsymbol{p}_{{b}_{k}}}$ can be rewritten as follows:
\begin{align}\label{eqn:delta p new}
	\delta _{ST}^{G}\boldsymbol{p}_{{b}_{k}}&= {^{G}\boldsymbol{p}_{{b}_{k}}} - {_{{b}_{k}}^{G}\boldsymbol{R}}{_{G}^{{b}_{k}}\boldsymbol{\widetilde{R}}}{^{G}\boldsymbol{\widetilde{p}}_{{b}_{k}}} \\
	&\approx - \delta ^{G}\boldsymbol{p}_{{b}_{k}} +[{^{G}\boldsymbol{\widetilde{p}}_{{b}_{k}}}\times]_{{b}_{k}}^{G}\boldsymbol{\phi}    \nonumber
\end{align}

The derivation process of the new position error differential equation is as:
\begin{align}\label{eqn:differential delta p new}
	\delta _{ST}^{G}\boldsymbol{\dot{p}}_{{b}_{k}}&= - \delta ^{G}\boldsymbol{\dot{p}}_{{b}_{k}} + [{^{G}\boldsymbol{\dot{\widetilde{p}}}_{{b}_{k}}}\times]_{{b}_{k}}^{G}\boldsymbol{\phi} + [{^{G}\boldsymbol{\widetilde{p}}_{{b}_{k}}}\times]_{{b}_{k}}^{G}\boldsymbol{\dot{\phi}}   \\
	&\approx -[^{G}\boldsymbol{p}_{{b}_{k}}\times][\boldsymbol{\omega}^{G}\times]{_{{b}_{k}}^{G}\boldsymbol{\phi}}+{\delta ^{G}\boldsymbol{v}_{{b}_{k}}} \nonumber \\ 
	&\quad - [^{G}\boldsymbol{p}_{{b}_{k}}\times]{_{{b}_{k}}^{G}\boldsymbol{R}}{\delta \boldsymbol{b}_{{g}_{k}}} - [^{G}\boldsymbol{p}_{{b}_{k}}\times]{_{{b}_{k}}^{G}\boldsymbol{R}}{w_{g}}										\nonumber
\end{align}
where $\boldsymbol{\omega}^{G}$ is the projection of the earth's angular velocity in $\{G\}$ frame.

The system error state model of IMU can be defined as:
\begin{align}\label{eqn:deltaxb process PO}
	\delta \boldsymbol{\dot{x}}_{b}&=\boldsymbol{F}_{b} \delta \boldsymbol{x}_{b}+\boldsymbol{G}_{b}\boldsymbol{w}_{b} \\ \boldsymbol{w}_{b}&=\begin{bmatrix} \boldsymbol{w}_{g} & \boldsymbol{w}_{a} & \boldsymbol{w}_{wg} & \boldsymbol{w}_{wa} \end{bmatrix} ^{T}
\end{align} 
where $\boldsymbol{w}_{b}$ is the system noise vector.  
$\boldsymbol{w}_{g}$ and $\boldsymbol{w}_{a}$ represent the measurement white noise of the gyro and accelerometer respectively; $\boldsymbol{w}_{wg}$ and $\boldsymbol{w}_{wa}$ represent the driven white noise of the gyro biases and accelerometer biases.

According to Eq. \eqref{eqn:differential delta p new}, the error-state transition matrix $\boldsymbol{F}_{b}$ and the input noise Jacobian matrix $\boldsymbol{G}_{b}$ can be represented as:
\begin{small}
\begin{align}\label{eqn:F PO}
	\boldsymbol{F}_{b}=
	\begin{bmatrix}
		- [\boldsymbol{\omega}^{G}\times] & \boldsymbol{0}_{3\times3} & \boldsymbol{0}_{3\times3} & -_{b}^{G}\boldsymbol{R} & \boldsymbol{0}_{3\times3} \\
		[\boldsymbol{g}^{G}\times]\\+[^{G}\boldsymbol{v}_{b}\times][\boldsymbol{\omega}^{G}\times] & -2 [\boldsymbol{\omega}^{G}\times] & \boldsymbol{0}_{3\times3} & -[^{G}\boldsymbol{v}_{b}\times]{_{b}^{G}\boldsymbol{R}} & -\boldsymbol{R}_{b}^{w} \\
		-[^{G}\boldsymbol{p}_{b}\times][\boldsymbol{\omega}^{G}\times] & \boldsymbol{I}_{3\times3} & \boldsymbol{0}_{3\times3} & -[^{G}\boldsymbol{p}_{b}\times]{_{b}^{G}\boldsymbol{R}} &\boldsymbol{0}_{3\times3} \\
		\boldsymbol{0}_{3\times3} & \boldsymbol{0}_{3\times3} & \boldsymbol{0}_{3\times3} & \boldsymbol{0}_{3\times3} &\boldsymbol{0}_{3\times3} \\
		\boldsymbol{0}_{3\times3} & \boldsymbol{0}_{3\times3} & \boldsymbol{0}_{3\times3} & \boldsymbol{0}_{3\times3} &\boldsymbol{0}_{3\times3}
	\end{bmatrix}
\end{align} 
\end{small}
\begin{align}\label{eqn:G PO}
	\boldsymbol{G}_{b}=
	\begin{bmatrix}
		-_{b}^{G}\boldsymbol{R} & \boldsymbol{0}_{3\times3} & \boldsymbol{0}_{3\times3} & \boldsymbol{0}_{3\times3} \\
		-[^{G}\boldsymbol{v}_{b}\times]{_{b}^{G}\boldsymbol{R}} & -{_{b}^{G}\boldsymbol{R}} & \boldsymbol{0}_{3\times3} & \boldsymbol{0}_{3\times3} \\
		-[^{G}\boldsymbol{p}_{b}\times]{_{b}^{G}\boldsymbol{R}} & \boldsymbol{0}_{3\times3} & \boldsymbol{0}_{3\times3} & \boldsymbol{0}_{3\times3} \\
		\boldsymbol{0}_{3\times3} & \boldsymbol{0}_{3\times3} & \boldsymbol{I}_{3\times3} & \boldsymbol{0}_{3\times3} \\
		\boldsymbol{0}_{3\times3} & \boldsymbol{0}_{3\times3} & \boldsymbol{0}_{3\times3} & \boldsymbol{I}_{3\times3} 
	\end{bmatrix}
\end{align}
where $\boldsymbol{g}^{G}$ is the gravity vector in $\{G\}$ frame.

The state and covariance augmentation is consistent with OpenVINS \cite{9196524} and will not be repeated here.
In addition, since we have modified the error state definition, the state update equation is modified as:
\begin{small}
\begin{align}\label{eqn:delta update}
	&\begin{bmatrix}
		{_{b}^{G}\boldsymbol{R}} & {^{G}\boldsymbol{v}_{b}} & {^{G}\boldsymbol{p}_{b}} \\
		\boldsymbol{0}_{1\times3} & 1 & 0 \\
		\boldsymbol{0}_{1\times3} & 0 & 1 \\
	\end{bmatrix} = \\ \nonumber 
	&\begin{bmatrix}
		[\boldsymbol{I}+{_{b}^{G}\boldsymbol{\phi}}\times ] & \delta{^{G}\boldsymbol{v}_{b}} & \delta{^{G}\boldsymbol{p}_{b}} \\
		\boldsymbol{0}_{1\times3} & -1 & 0 \\
		\boldsymbol{0}_{1\times3} & 0 & -1 \\
	\end{bmatrix}
	\begin{bmatrix}
		{_{b}^{G}\boldsymbol{\widetilde{R}}} & {^{G}\boldsymbol{\widetilde{v}}_{b}} & {^{G}\boldsymbol{\widetilde{p}}_{b}} \\
		\boldsymbol{0}_{1\times3} & -1 & 0 \\
		\boldsymbol{0}_{1\times3} & 0 & -1 \\
	\end{bmatrix}
\end{align}
\end{small}

\subsection{Measurement Model Based on PO Visual description}\label{sec:Measurement Model Based on PO Visual description}
Since the visual residual model is related to 3D features, the accuracy of MSCKF-based VIO is limited by the 3D reconstruction process. 
With reference to PO theory \cite{WOS:000458768000003,WOS:000899419900005} , we reconstruct the measurement model, which is represented only by pixel coordinates and relative pose.

\subsubsection{PO Description of Multiple View Geometry}\label{sec:Fundamentals of Pose-only Theory}
Assuming a 3D feature $^{G}\boldsymbol{p}_{f}=\begin{bmatrix} ^{G}{X}_{f} & ^{G}{Y}_{f} & ^{G}{Z}_{f} \end{bmatrix}$ observed in $n$ images, its normalized coordinate in the $i$-th image is $\boldsymbol{p}_{{C}_{i}}=\begin{bmatrix} {X}_{{C}_{i}} & {Y}_{{C}_{i}} & 1 \end{bmatrix}$ ($i=1,...n$).

Defining the global rotation and global translation of camera in the $i$-th image as $_{G}^{{C}_{i}}\boldsymbol{R}$ and $^{G}\boldsymbol{t}_{{C}_{i}}$, then the projection equation of 3D feature $^{G}\boldsymbol{p}_{f}$ can be expressed as:
\begin{align}\label{eqn:feature projection}	
	\boldsymbol{p}_{{C}_{i}}=\frac{1}{^{{C}_{i}}{Z}_{f}}{^{{C}_{i}}{p}_{f}}=\frac{1}{{Z}_{f}^{{C}_{i}}}{_{G}^{{C}_{i}}\boldsymbol{R}}(^{G}\boldsymbol{p}_{f}-^{G}\boldsymbol{t}_{{C}_{i}})
\end{align}
where $^{{C}_{i}}\boldsymbol{p}_{f}=\begin{bmatrix} ^{{C}_{i}}{X}_{f} & ^{{C}_{i}}{Y}_{f} & ^{{C}_{i}}{Z}_{f} \end{bmatrix}$ represents the position of 3D feature $^{G}\boldsymbol{p}_{f}$ in the $i$-th image and 
$^{{C}_{i}}{Z}_{f} \textgreater 0$ represents the depth of this feature.

As shown in \cite{WOS:000899419900005}, when a 3D feature is observed in $n$ images, traditional multi-view geometry descriptions can be equivalently expressed in a pose-only form:
\begin{small}
	\begin{align}\label{eqn:constraint mulimage}	
		D(j,k)=\left\{{d}_{i}^{(j,i)}\boldsymbol{p}_{{C}_{i}}={d}_{j}^{(j,k)}{_{{C}_{j}}^{{C}_{i}}\boldsymbol{R}}\boldsymbol{p}_{{C}_{j}}+{^{{C}_{i}}\boldsymbol{t}_{{C}_{j}}}|1 \leq i \leq n, i \neq j\right\}
	\end{align}
\end{small}

The $j$-th and $k$-th images represent the left and right base views of the constraint. The suggested base view selection method can be expressed as \cite{WOS:000899419900005}:
\begin{align}\label{eqn:baseframe sel}	
	(j,k)=\mathop{argmax}\limits_{1 \leq j,k \leq n} \left\{ \boldsymbol{\theta}_{j,k} \right\} 
\end{align}

In traditional multi-view geometry descriptions, assuming $\boldsymbol{p}_{{c}_{i}}^{(l)}$ is the normalized coordinate of feature $l$ in $i$-th image, the reprojection error can be defined as:
\begin{align}\label{eqn:repro BA}
	\boldsymbol{r}_{{C}_{i}}^{(l)}=\boldsymbol{\widetilde{p}}_{{C}_{i}}^{(l)}-\boldsymbol{p}_{{C}_{i}}^{(l)}=\frac{{^{{C}_{i}}\boldsymbol{p}_{{f}_{l}}^{(BA)}}}{{{e}_{3}^{T}}{^{{C}_{i}}\boldsymbol{p}_{{f}_{l}}^{(BA)}}}-\boldsymbol{p}_{{C}_{i}}^{(l)}
\end{align}
where ${^{{C}_{i}}\boldsymbol{p}_{{f}_{l}}^{(BA)}}={_{G}^{{C}_{i}}\boldsymbol{R}}(^{G}\boldsymbol{p}_{{f}_{l}}^{(BA)}-{^{G}\boldsymbol{t}_{{C}_{i}}})$ and ${e}_{3}^{T}=\begin{bmatrix} 0 & 0 & 1 \end{bmatrix}$. 
$^{G}\boldsymbol{p}_{{f}_{l}}^{(BA)}$ can be estimated through a nonlinear optimization process that minimizes reprojection errors, mainly achieved through triangulation measurement and Bundle Adjustment (BA) \cite{10.5555/1888028.1888032}.

After using PO multi-view geometry description in \eqref{eqn:constraint mulimage}, the reprojection error can be redefined as:

\begin{align}\label{eqn:repro PA}
	\boldsymbol{r}_{{c}_{i}}^{(l)}=\boldsymbol{\widetilde{p}}_{{c}_{i}}^{(l)}-\boldsymbol{p}_{{c}_{i}}^{(l)}=\frac{{^{{C}_{i}}\boldsymbol{p}_{{f}_{l}}^{(PO)}}}{{{e}_{3}^{T}}{^{{C}_{i}}\boldsymbol{p}_{{f}_{l}}^{(PO)}}}-\boldsymbol{p}_{{c}_{i}}^{(l)}
\end{align}

where $^{{C}_{i}}\boldsymbol{p}_{{f}_{l}}^{(PO)}$ can be directly obtained through the relative camera pose and 2D features as:

\begin{small}
\begin{align}\label{eqn:3Dpos PO}
	^{{C}_{i}}\boldsymbol{p}_{{f}_{l}}^{(PO)}= ||[{^{{C}_{k}}\boldsymbol{t}_{{C}_{j}}}\times]\boldsymbol{p}_{{c}_{k}}||{_{{C}_{j}}^{{C}_{i}}\boldsymbol{R}}\boldsymbol{p}_{{c}_{j}}+||[{\boldsymbol{p}_{{c}_{k}}}\times]{_{{C}_{j}}^{{C}_{i}}\boldsymbol{R}}{\boldsymbol{p}_{{c}_{j}}}||{^{{c}_{i}}\boldsymbol{t}_{{c}_{j}}}
\end{align}
\end{small}
where the $j$-th and $k$-th images represent the left and right base views respectively.

Compared to \eqref{eqn:repro BA}, the reprojection error defined in \eqref{eqn:repro PA} is independent of 3D features, and can eliminate the 3D reconstruction process.

\subsubsection{PO-Based State Update}\label{sec:Measurement Model}

In MSCKF-based VIO, the reprojection error is related to the camera poses and 3D features, but the system state does not contain 3D features.
Therefore, it is necessary to use the left nullspace to restore the standard EKF format before filtering updates.

In \eqref{eqn:repro PA}, the reprojection error $\boldsymbol{r}_{{C}_{i}}^{(l)}$ is independent of 3D features, then it can be rewritten as:

\begin{align}\label{eqn:reproject PO}
	\boldsymbol{r}_{{C}_{i}}^{(l)} \approx \boldsymbol{H}_{{x}_{i}}^{(l)} \delta \boldsymbol{x} + \boldsymbol{n}_{i}^{(l)}
\end{align}

Compared to MSCKF \cite{4209642}, \eqref{eqn:reproject PO} is the standard EKF measurement model and does not require the left nullspace.

$\boldsymbol{H}_{{x}_{i}}^{(l)}$ is the Jacobian matrix of $\boldsymbol{r}_{{c}_{i}}^{(l)}$ relative to $\delta \boldsymbol{x}$, and the derivation process is as follows:

\begin{equation}\label{eqn:Hxi PO}
	\begin{aligned}
		\boldsymbol{H}_{{x}_{i}}^{(l)}&=\frac{\partial\boldsymbol{r}_{{C}_{i}}^{(l)}}{\partial\delta \boldsymbol{x}}
		=\frac{\partial\boldsymbol{r}_{{C}_{i}}^{(l)}}{\partial{^{{C}_{i}}\boldsymbol{p}_{{f}_{l}}^{(PO)}}}
		\frac{\partial{^{{C}_{i}}\boldsymbol{p}_{{f}_{l}}^{(PO)}}}{\partial\delta \boldsymbol{x}} \\
		&=\begin{bmatrix}
			\frac{1}{{e}_{3}^{T}{^{{C}_{i}}\boldsymbol{p}_{{f}_{l}}^{(PO)}}} & 0 & -\frac{{e}_{1}^{T}{^{{C}_{i}}\boldsymbol{p}_{{f}_{l}}^{(PO)}}}{{({e}_{3}^{T}{^{{C}_{i}}\boldsymbol{p}_{{f}_{l}}^{(PO)}})}^2} \\
			0 & \frac{1}{{e}_{3}^{T}{^{{C}_{i}}\boldsymbol{p}_{{f}_{l}}^{(PO)}}} & -\frac{{e}_{2}^{T}{^{{C}_{i}}\boldsymbol{p}_{{f}_{l}}^{(PO)}}}{{({e}_{3}^{T}{^{{C}_{i}}\boldsymbol{p}_{{f}_{l}}^{(PO)}})}^2}
		\end{bmatrix} \\
		&~~\cdot\begin{bmatrix}
			\boldsymbol{0}_{3\times15} & \frac{\partial{^{{C}_{i}}\boldsymbol{p}_{{f}_{l}}^{(PO)}}}{\partial{_{{b}_{m}}^{G}\boldsymbol{\phi}}} & \frac{\partial{^{{C}_{i}}\boldsymbol{p}_{{f}_{l}}^{(PO)}}}{\partial\delta{^{G}\boldsymbol{p}_{{b}_{m}}}} & ...
		\end{bmatrix}
	\end{aligned}
\end{equation}
where ${e}_{1}^{T}=\begin{bmatrix} 1 & 0 & 0 \end{bmatrix}$, ${e}_{2}^{T}=\begin{bmatrix} 0 & 1 & 0 \end{bmatrix}$, and ${e}_{3}^{T}=\begin{bmatrix} 0 & 0 & 1 \end{bmatrix}$.
The subscript $m$ represents the ID number of historical clone and satisfy $1 \leq m \leq n$.

Notably, since only imu poses are saved in the historical clone, the camera poses at the corresponding time are obtained by ${_{G}^{C}\boldsymbol{R}} = {_{b}^{C}\boldsymbol{R}}{_{G}^{b}\boldsymbol{R}} $ and $^{G}\boldsymbol{p}_{C} = {^{G}\boldsymbol{p}_{b}} + {_{b}^{G}\boldsymbol{R}}{^{b}\boldsymbol{p}_{C}}$. Where ${^{b}\boldsymbol{p}_{C}}$ and ${_{b}^{C}\boldsymbol{R}}$ represent the transformation relationship between the body frame $\{b\}$ and the camera frame $\{C\}$. 

The partial derivative of ${^{{C}_{i}}\boldsymbol{p}_{{f}_{l}}^{(PO)}}$ with respect to ${_{{b}_{m}}^{G}\boldsymbol{\phi}}$ and ${\delta{^{G}\boldsymbol{p}_{{b}_{m}}}}$ can be respectively expressed as:
\begin{footnotesize}
\begin{equation}\label{eqn:Hxi phi PO}
	\begin{aligned}
		\frac{\partial{^{{C}_{i}}\boldsymbol{p}_{{f}_{l}}^{(PO)}}}{\partial{_{{b}_{m}}^{G}\boldsymbol{\phi}}}&=
		\frac{\partial(||[{^{{C}_{k}}\boldsymbol{t}_{{C}_{j}}}\times]\boldsymbol{p}_{{C}_{k}}||
			{_{{C}_{j}}^{{C}_{i}}\boldsymbol{R}}\boldsymbol{p}_{{C}_{j}}
			+||[\boldsymbol{p}_{{C}_{k}}\times]{_{{C}_{j}}^{{C}_{k}}\boldsymbol{R}}\boldsymbol{p}_{{C}_{j}}||\boldsymbol{^{{C}_{i}}\boldsymbol{t}_{{C}_{j}}})}{\partial{_{{b}_{m}}^{G}\boldsymbol{\phi}}} 
	\end{aligned}
\end{equation}
\end{footnotesize}

\begin{footnotesize}
\begin{equation}\label{eqn:Hxi pos PO}
	\begin{aligned}
		\frac{\partial{^{{C}_{i}}\boldsymbol{p}_{{f}_{l}}^{(PO)}}}{\partial\delta{^{G}\boldsymbol{p}_{{b}_{m}}}}&=
		\frac{\partial(||[{^{{C}_{k}}\boldsymbol{t}_{{C}_{j}}}\times]\boldsymbol{p}_{{C}_{k}}||
			{_{{C}_{j}}^{{C}_{i}}\boldsymbol{R}}\boldsymbol{p}_{{C}_{j}}
			+||[\boldsymbol{p}_{{C}_{k}}\times]{_{{C}_{j}}^{{C}_{k}}\boldsymbol{R}}\boldsymbol{p}_{{C}_{j}}||{^{{C}_{i}}\boldsymbol{t}_{{C}_{j}}})}{\partial\delta{^{G}\boldsymbol{p}_{{b}_{m}}}} 
	\end{aligned}
\end{equation}
\end{footnotesize}

As shown in \eqref{eqn:reproject PO} to \eqref{eqn:Hxi pos PO}, the new measurement model is represented only by pixel coordinates and system pose, and is completely decoupled from the 3D features, thus avoiding the impact of inaccurate 3D reconstruction processes.

\section{Observability Analysis}\label{sec:Observability Analysis}
Observability is an important metric for consistency analysis of filter-based VIO systems, and the observability matrix for the linearized system can  be construct following \cite{6605544,doi:10.1177/0278364909353640}:
\begin{align}\label{eqn:Observability matrix}
	\boldsymbol{O} = 
        \begin{bmatrix}
		\boldsymbol{O}_{0} \\
		\boldsymbol{O}_{1} \\
		\vdots \\
		\boldsymbol{O}_{k}
	\end{bmatrix} =
	\begin{bmatrix}
		\boldsymbol{H}_{0} \\
		\boldsymbol{H}_{1}{\boldsymbol{\Phi}_{1,0}} \\
		\vdots \\
		\boldsymbol{H}_{k}{\boldsymbol{\Phi}_{k,0}}
	\end{bmatrix}
\end{align}
where ${\boldsymbol{\Phi}_{k,0}}={\boldsymbol{\Phi}_{k,k-1}} \cdots {\boldsymbol{\Phi}_{1,0}}$ is the state transition matrix with respect to the time interval $[0, k]$, $\boldsymbol{H}_{k}$ is the measured Jacobian matrix of the system at time $k$.

If $\boldsymbol{O}$ is full column rank, the system is fully observable. However, the VIO system is partially observable with a nullspace $\boldsymbol{N}$ satisfying \cite{6605544,doi:10.1177/0278364909353640}:
\begin{align}\label{eqn:O N}
	{\boldsymbol{O}_{k}}{\boldsymbol{N}_{k}} = \boldsymbol{H}_{k}{\boldsymbol{\Phi}_{k,0}}{\boldsymbol{N}_{k}} = \boldsymbol{0}
\end{align}
where ${\boldsymbol{N}}$ describes the state unobservable subspace that can not be recovered with measurements.

In this section, we start with the unobservable subspace of filter-based VIO systems, first perform observability analysis of several traditional multi-view geometry based VIO systems, and then discuss the effect of the introduction of pose-only visual description on the system's observability and whether there is consistency improvement.

\subsection{Multi-View geometry Based VIO Systems}\label{sec:Multi-View geometry based VIO Systems}
Due to the measurement model format of MSCKF-based VIO\cite{9196524,WOS:000411059400002,WOS:000424646100016,indelman2012real,4209642}, observability analysis is difficult to perform directly. It is proved in \cite{doi:10.1177/0278364913481251} that the observability analysis of EKF-SLAM and MSCKF is equivalent. Therefore, the unobservable subspace of several algorithms are calculated on the basis of the linearized model of EKF-SLAM in this subsection. 

Assuming that a single visual feature is observed at time $k$, the state vector is defined as:
\begin{align}
	{\boldsymbol{x}} = \begin{bmatrix} _{{b}_{k}}^{G}\boldsymbol{q}^{T} & ^{G}\boldsymbol{v}_{{b}_{k}}^{T} & ^{G}\boldsymbol{p}_{{b}_{k}}^{T} & \boldsymbol{b}_{{g}_{k}}^{T} & \boldsymbol{b}_{{a}_{k}}^{T} & ^{G}\boldsymbol{p}_{f}^{T} \end{bmatrix} ^{T}
\end{align}

From \cite{6605544}, the unobservable subspace of Std-EKF based VIO systems \cite{9196524,4209642} can be expressed as:
\begin{equation}\label{eqn:N_k_stdekfVINS}
	\begin{aligned}
		{\boldsymbol{N}_{k}^{Std}} = \begin{bmatrix}
			{\boldsymbol{0}_{3\times3}} & {\boldsymbol{g}^{G}} \\
			{\boldsymbol{0}_{3\times3}} & -[{^{G}{\boldsymbol{v}}_{b_0}} \times]{\boldsymbol{g}^{G}} \\
			{\boldsymbol{I}_{3\times3}} & -[{^{G}{\boldsymbol{p}}_{b_0}} \times]{\boldsymbol{g}^{G}} \\
			{\boldsymbol{0}_{3\times3}} & {\boldsymbol{0}_{3\times1}} \\
			{\boldsymbol{0}_{3\times3}} & {\boldsymbol{0}_{3\times3}} \\
			{\boldsymbol{I}_{3\times3}} & [{^{G}{\boldsymbol{p}}_{f}} \times]{\boldsymbol{g}^{G}} 
		\end{bmatrix}
	\end{aligned}
\end{equation}

From \cite{6605544}, the unobservable subspace in \eqref{eqn:N_k_stdekfVINS} is susceptible to system linearization change, resulting in spurious information inflation to the unobservable direction causing inconsistency problems.

To solve the inconsistent problem, \cite{8205965} proposed the right invariant (RL)-VINS, which adopts right invariant errors for VINS states except for IMU biases. The unobservable subspace of RL-VINS can be computed as\cite{yang2022decoupled}:
\begin{equation}\label{eqn:N_k_RIVINS}
	\begin{aligned}
		{\boldsymbol{N}_{k}^{RI}} = \begin{bmatrix}
			{\boldsymbol{0}_{3\times3}} & {\boldsymbol{g}^{G}} \\
			{\boldsymbol{0}_{3\times3}} & {\boldsymbol{0}_{3\times1}} \\
			{\boldsymbol{I}_{3\times3}} & {\boldsymbol{0}_{3\times1}} \\
			{\boldsymbol{0}_{3\times3}} & {\boldsymbol{0}_{3\times1}} \\
			{\boldsymbol{0}_{3\times3}} & {\boldsymbol{0}_{3\times3}} \\
			{\boldsymbol{I}_{3\times3}} & {\boldsymbol{0}_{3\times1}} 
		\end{bmatrix}
	\end{aligned}
\end{equation}

As shown in \eqref{eqn:N_k_RIVINS}, the null space of RI-VINS is unrelated to the state vector and can directly satisfy the unobservable properties of VINS without Jacobian modifications.
However, the disadvantage of RI-VINS is that the visual features participate in state covariance propagation in Lie-Group representation.
Assuming that the state vector contains $m$ features, the computation cost of covariance propagation for Std-ekf based VIO systems is only around $\textbf{O}(m)$ but can rise to $\textbf{O}(m^{2})$ for RI-VINS\cite{yang2022decoupled}, which significantly increases the computational cost.

Different from RI-VINS, DST-EKF based VIO only carries out state transformation for velocity and position errors without modifying the visual features state representation so that the unobservable subspace can be expressed as:
\begin{equation}\label{eqn:N_k_DRIVINS}
	\begin{aligned}
		{\boldsymbol{N}_{k}^{DST}} = \begin{bmatrix}
			{\boldsymbol{0}_{3\times3}} & {\boldsymbol{g}^{G}} \\
			{\boldsymbol{0}_{3\times3}} & {\boldsymbol{0}_{3\times1}} \\
			{\boldsymbol{I}_{3\times3}} & {\boldsymbol{0}_{3\times1}} \\
			{\boldsymbol{0}_{3\times3}} & {\boldsymbol{0}_{3\times1}} \\
			{\boldsymbol{0}_{3\times3}} & {\boldsymbol{0}_{3\times3}} \\
			{\boldsymbol{I}_{3\times3}} & -[{^{G}{\boldsymbol{p}}_{f}} \times]{\boldsymbol{g}^{G}} 
		\end{bmatrix}
	\end{aligned}
\end{equation}

From \eqref{eqn:N_k_stdekfVINS} to \eqref{eqn:N_k_DRIVINS}, DST-EKF based VIO has better consistency than Std-ekf based VIO, and does not suffer from the computational costs associated with visual feature amplification compared to RI-VINS.
However, the feature estimation will affect its unobservable subspace,, and the system will suffer from the inconsistency caused by the feature linearization error when the feature estimation is not accurate.

In addition, similar work has been done with Dri-VINS\cite{yang2022decoupled}, which reduces the computational cost by decoupling features from Lie group representations based on RI-VINS.
Dri-VINS and DST-EKF based VIO have similar observability but are also plagued by feature linearization errors. Therefore, FEJ method \cite{FEJ} is introduced in \cite{yang2022decoupled} to maintain the system's consistency.

Since multi-view geometry based VIO is associated with 3D features, when 3D features are not included in the state vectors, the linearization error introduced by feature triangulation will destroy the system's consistency, but adding 3D features to the state covariance propagation will lead to a surge in computing costs and limit the application of the system under different conditions.

\subsection{Pose-Only Visual description Based VIO Systems}\label{sec:Pose-Only Visual description Based VIO Systems}
Unlike multi-view geometry based VIO, PO based VIO is decoupled from 3D features, so 3D features are not added to the state vector in observability analysis.
Assuming that a single visual feature is observed in the images at three moments, and the images at time $\alpha$ and time $\beta$ are taken as the left/right-base views, the state vectors of PO-EKF and SP-VIO at time $k$ can be defined as:
\begin{equation}\label{eqn:x observation_po}
	\begin{aligned}
		\boldsymbol{x}&=\begin{bmatrix} \boldsymbol{x}_{b}^{T} & \boldsymbol{x}_{c}^{T} \end{bmatrix} ^{T} \\
		\boldsymbol{x}_{b}&=\begin{bmatrix} _{{b}_{k}}^{G}\boldsymbol{q}^{T} & ^{G}\boldsymbol{v}_{{b}_{k}}^{T} & ^{G}\boldsymbol{p}_{{b}_{k}}^{T} & \boldsymbol{b}_{{g}_{k}}^{T} & \boldsymbol{b}_{{a}_{k}}^{T} \end{bmatrix} ^{T} \\
		\boldsymbol{x}_{c}&=\begin{bmatrix} _{{b}_{\alpha}}^{G}\boldsymbol{q}^{T} & ^{G}\boldsymbol{p}_{{b}_{\alpha}}^{T} & _{{b}_{\beta}}^{G}\boldsymbol{q}^{T} & ^{G}\boldsymbol{p}_{{b}_{\beta}}^{T}  \end{bmatrix} ^{T} 
	\end{aligned}
\end{equation}

From \eqref{eqn:Hxi PO} and \eqref{eqn:x observation_po}, the Jacobian matrix of PO-EKF and SP-VIO at time $k$ is defined as:
\begin{equation}\label{eqn:H_x_ob_ham}
	\begin{aligned}
		\boldsymbol{H}_{x_k} =  \boldsymbol{H}_{p_f}
            [ {\boldsymbol{H}_{{\phi}_{k}}} \quad {\boldsymbol{0}_{3\times3}} \quad {\boldsymbol{H}_{{p}_{k}}} \quad {\boldsymbol{0}_{3\times6}} \quad \dots \\
             {\boldsymbol{H}_{{\phi}_{\alpha}}} \quad {\boldsymbol{H}_{{p}_{\alpha}}} \quad {\boldsymbol{H}_{{\phi}_{\beta}}} \quad {\boldsymbol{H}_{{p}_{\beta}}} ]
	\end{aligned}
\end{equation}

Combine \cite{6605544} and \eqref{eqn:H_x_ob_ham}, the unobservable subspace of PO-EKF can be computed as:
\begin{equation}\label{eqn:N_k_poekf}
	\begin{aligned}
		{\boldsymbol{N}_{k}^{PO}} = \begin{bmatrix}
			{\boldsymbol{0}_{3\times3}} & \left({^{G}\boldsymbol{p}_{b_k}}-{^{G}\boldsymbol{p}_{b_{\alpha}}}  -{^{G}\boldsymbol{v}_{b_{\alpha}}} + {\boldsymbol{g}^{G}}\right) \\
			{\boldsymbol{0}_{3\times3}} & {\boldsymbol{0}_{3\times1}} \\
			{\boldsymbol{I}_{3\times3}} & -{[\left( {^{G}\boldsymbol{p}_{b_k}} - {^{G}\boldsymbol{p}_{b_\alpha}} \right) \times]} \\ 
            &\left({^{G}\boldsymbol{p}_{b_k}}-{^{G}\boldsymbol{p}_{b_{\alpha}}}  -{^{G}\boldsymbol{v}_{b_{\alpha}}} + {\boldsymbol{g}^{G}}\right) \\
			{\boldsymbol{0}_{3\times3}} & {\boldsymbol{0}_{3\times1}} \\
			{\boldsymbol{0}_{3\times3}} & {\boldsymbol{0}_{3\times1}} \\
			{\boldsymbol{0}_{3\times3}} & \left({^{G}\boldsymbol{p}_{b_k}}-{^{G}\boldsymbol{p}_{b_{\alpha}}}  -{^{G}\boldsymbol{v}_{b_{\alpha}}} + {\boldsymbol{g}^{G}}\right) \\
			{\boldsymbol{I}_{3\times3}} & {\boldsymbol{0}_{3\times1}} \\
			{\boldsymbol{0}_{3\times3}} & {\boldsymbol{0}_{3\times1}} \\
			{\boldsymbol{I}_{3\times3}} & {\boldsymbol{0}_{3\times1}} 
		\end{bmatrix}
	\end{aligned}
\end{equation}

Similar to \eqref{eqn:N_k_stdekfVINS}, although the unobservable subspace of PO-EKF is decouples from the visual features, it is also susceptible to linearization errors, resulting in inconsistency problems.

After replacing Std-EKF with DST-EKF, the unobservable subspace of SP-VIO can be expressed as:
\begin{equation}\label{eqn:N_k_spvio}
	\begin{aligned}
		{\boldsymbol{N}_{k}^{SP}}  = 
		\begin{bmatrix}
			{\boldsymbol{0}_{3\times3}} & {\boldsymbol{g}^{G}} \\
			{\boldsymbol{0}_{3\times3}} & {\boldsymbol{0}_{3\times1}} \\
			{\boldsymbol{I}_{3\times3}} & -{[\left( {^{G}\boldsymbol{p}_{b_k}} - {^{G}\boldsymbol{p}_{b_\alpha}} \right) \times]}{\boldsymbol{g}^{G}} \\
			{\boldsymbol{0}_{3\times3}} & {\boldsymbol{0}_{3\times1}} \\
			{\boldsymbol{0}_{3\times3}} & {\boldsymbol{0}_{3\times1}} \\
			{\boldsymbol{0}_{3\times3}} &  {\boldsymbol{g}^{G}} \\
			{\boldsymbol{I}_{3\times3}} & {\boldsymbol{0}_{3\times1}} \\
			{\boldsymbol{0}_{3\times3}} & {\boldsymbol{0}_{3\times1}} \\
			{\boldsymbol{I}_{3\times3}} & {\boldsymbol{0}_{3\times1}}
		\end{bmatrix}
	\end{aligned}
\end{equation}

As shown in \eqref{eqn:N_k_spvio}, the unobservable subspace of SP-VIO is less affected by system linearization changes than PO-EKF.
Meanwhile, compared with the multi-view geometry based VIO systems mentioned in \secref{sec:Multi-View geometry based VIO Systems}, SP-VIO can achieve better consistency while maintaining higher computational efficiency after decoupling from 3D visual features.

\section{Enhanced SP-VIO Combined with DST-RTS Correction Strategy}\label{sec:Barebones SP-VIO}
Due to camera's sensitivity, visual tracking failure is easy to occur in challenging environments, which results in discontinuous visual measurements and affects VINS's localization accuracy.

In this section, we first analyze the coping strategy and output trajectory of the mainstream VIO algorithm under visual deprived conditions, 
and then propose an enhanced RTS correction strategy \cite{rts} combined with the DST-EKF proposed in \secref{sec:Barebones SP-VIO} to improve the robustness of SP-VIO under visual deprivation conditions.

\subsection{Mainstream VIO Algorithms}\label{sec:Mainstream VIO Algorithms}
\begin{figure}[htbp]
	\centering
	\subfigure[] { \begin{minipage}{4cm} \centering \includegraphics [height=3.2cm]{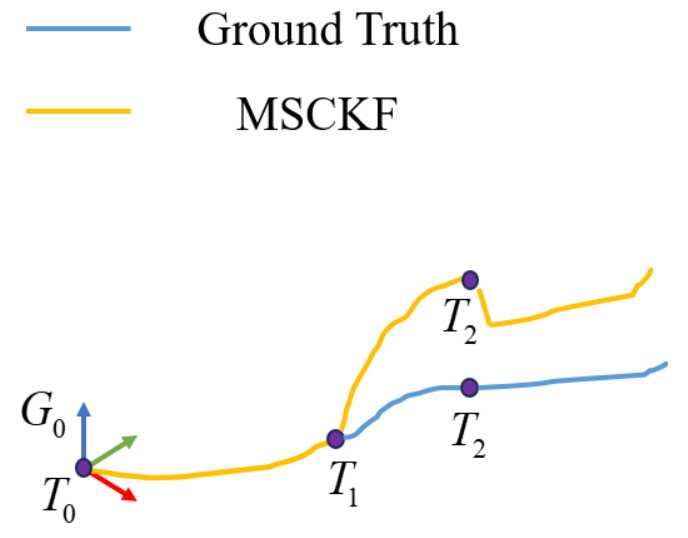}\label{fig:MSCKF1} \end{minipage} } 
	\subfigure[] { \begin{minipage}{4cm} \centering \includegraphics [height=3.2cm]{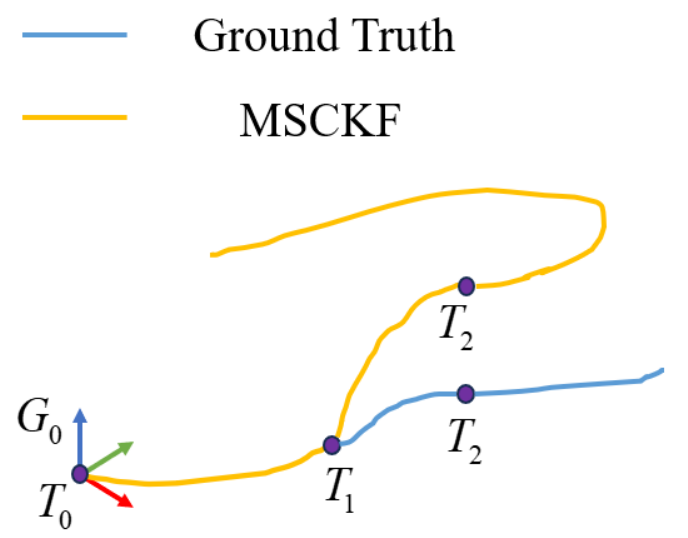}\label{fig:MSCKF2} \end{minipage} }
	\caption{
		The output trajectory of MSCKF under visual deprived condition.
		(a) estimated state reconvergence.
		(b) estimated state divergence.
	}
	\label{fig:MSCKF visual deprived}
\end{figure}

During visual deprivation, MSCKF \cite{4209642} performs an inertial navigation solution and updates the error covariance matrix.
As the linearization error of inertial navigation accumulates, the error covariance increases gradually.
When the observation is recovered, MSCKF will correct the estimated state to the direction of visual measurement, resulting in a jump in the estimated result, as shown in \figref{fig:MSCKF1}. If the visual measurement is more accurate, the jump direction will somewhat reduce the current error.

However, when the cumulative error during visual deprivation is large, the state estimation cannot converge after visual measurement recovery, resulting in the failure of filtering correction and divergence of navigation results, \figref{fig:MSCKF2}.

Consistent with MSCKF, VINS-Mono \cite{WOS:000442341000003} is also first settled based on inertial navigation when the visual observation is interrupted.
After the observation is restored, if the state estimation results can converge again, the navigation results will continue to be output under the original global frame $\{{G}_{0}\}$, as shown in \figref{fig:VINS1}:

\begin{figure}[htbp]
	\centering
	\subfigure[] { \begin{minipage}{4cm} \centering \includegraphics [height=3.2cm]{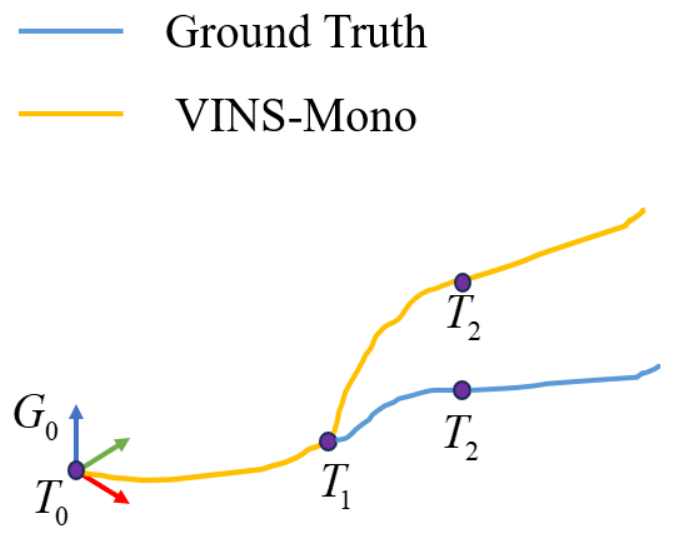}\label{fig:VINS1} \end{minipage} } 
	\subfigure[] { \begin{minipage}{4cm} \centering \includegraphics [height=3.2cm]{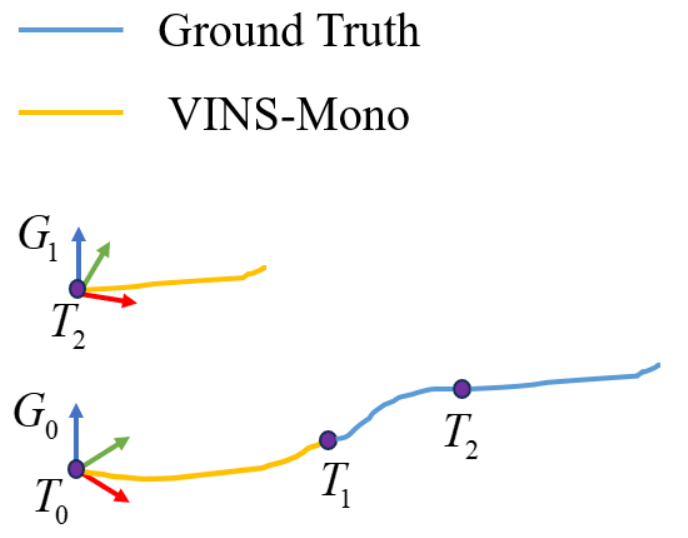}\label{fig:VINS2} \end{minipage} }
	\caption{
		The output trajectory of VINS-Mono under visual deprived condition.
		(a) estimated state reconvergence.
		(b) estimated state divergence.
	}
	\label{fig:VINS-Mono visual deprived}
\end{figure}

Since the construction of the nonlinear optimization function in VINS-Mono \cite{WOS:000442341000003} is mainly based on the relative relationship between IMU and visual measurements, the estimated results can only correct the local relative error and cannot affect the cumulative error during the visual deprivation, as shown in \figref{fig:VINS1}.
Also for this reason, VINS-Mono does not show the estimated jump in \figref{fig:MSCKF2}.

In the other case, if the inertial navigation solution error is large and the optimization state cannot be convergent, VINS-Mono will enter the re-initialization process, establishes a new global frame $\{{G}_{1}\}$, and outputs the correct navigation results again \cite{WOS:000442341000003}, as shown in \figref{fig:VINS2}.

\begin{figure}[htbp]
	\centering
	\includegraphics [width=3.5in]{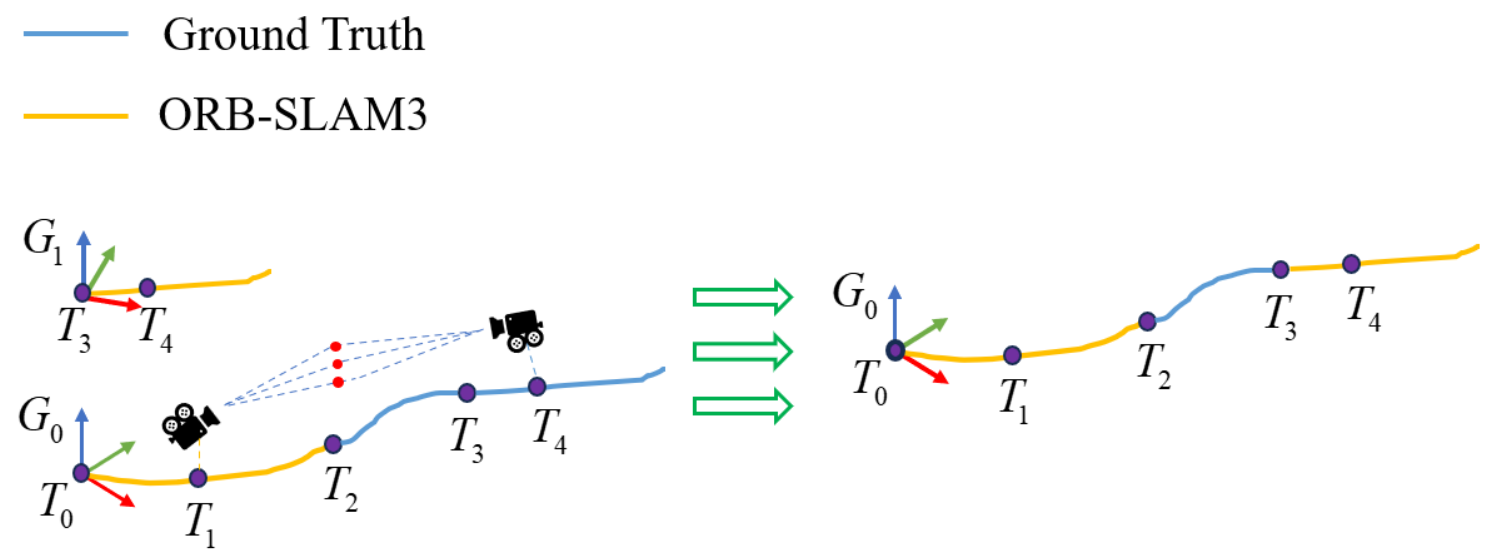} 
	\caption{
		The output trajectory of ORB-SLAM3 under visual deprived condition, with visual relocalization.
	}
	\label{fig:ORB-SLAM3 visual deprived}
\end{figure}

Also as the optimization-based VIO, ORB-SLAM3 \cite{WOS:000725804900006} handles visual deprived conditions like \figref{fig:VINS-Mono visual deprived}. 
Notably, due to the introduction of the multi-map system ORB-SLAM-Atlas \cite{WOS:000725804900006} , it can save the map before the observation interruption and re-establish the submap after the observation has been restored.
Therefore, when the visual loop closure is satisfied, ORB-SLAM3 fuses the two sub-maps to achieve the unity of the global frame $\{{G}_{0}\}$ , as shown in \figref{fig:ORB-SLAM3 visual deprived}.

In summary, it can be found that MSCKF and VINS-Mono lack the ability to correct the accumulated errors during visual deprivation, while ORB-SLAM3 relies on previous positioning scenes.
Therefore, the study of a correction strategy that does not rely on the visual loop closure will help improve the positioning performance of VIO systems under visual deprived conditions.

\subsection{DST-RTS Backtracking Smoothing}\label{sec:DST-RTS Backtracking Smoothing}
As shown in \cite{van2020invariant}, the propagation process of linearized systems with matrix Lie group element will be independent of the state estimate, so the RTS smoothing method with Lie group representation can better deal with the motion trajectory on poor initial state estimation.
From \secref{sec:Observability Analysis}, DST-EKF has similar propagation characteristics with the filter algorithms represented by matrix Lie group.
Therefore, DST-EKF is adopted in this section to reconstruct RTS smoothing to deal with the motion trajectory under visual deprivation condition.

The specific steps to enhanced RTS backtracking smoothing are as follows:

(1) Visual observation interruption: inertial navigation solution and state prediction;

(2) Visual observation recovery: integrated navigation with the velocity provided by VIO system;

(3) State estimation convergence: backtracking correction.

Notably, (1), (2) remarked as 'DST-EKF-forward', and (3) remarked as 'DST-EKF-RTS'. The timing chart of the proposed method is shown in \figref{fig:RTSintegram T}:
 
\begin{figure}[htbp]
	\centering
	\includegraphics [width=3.4in]{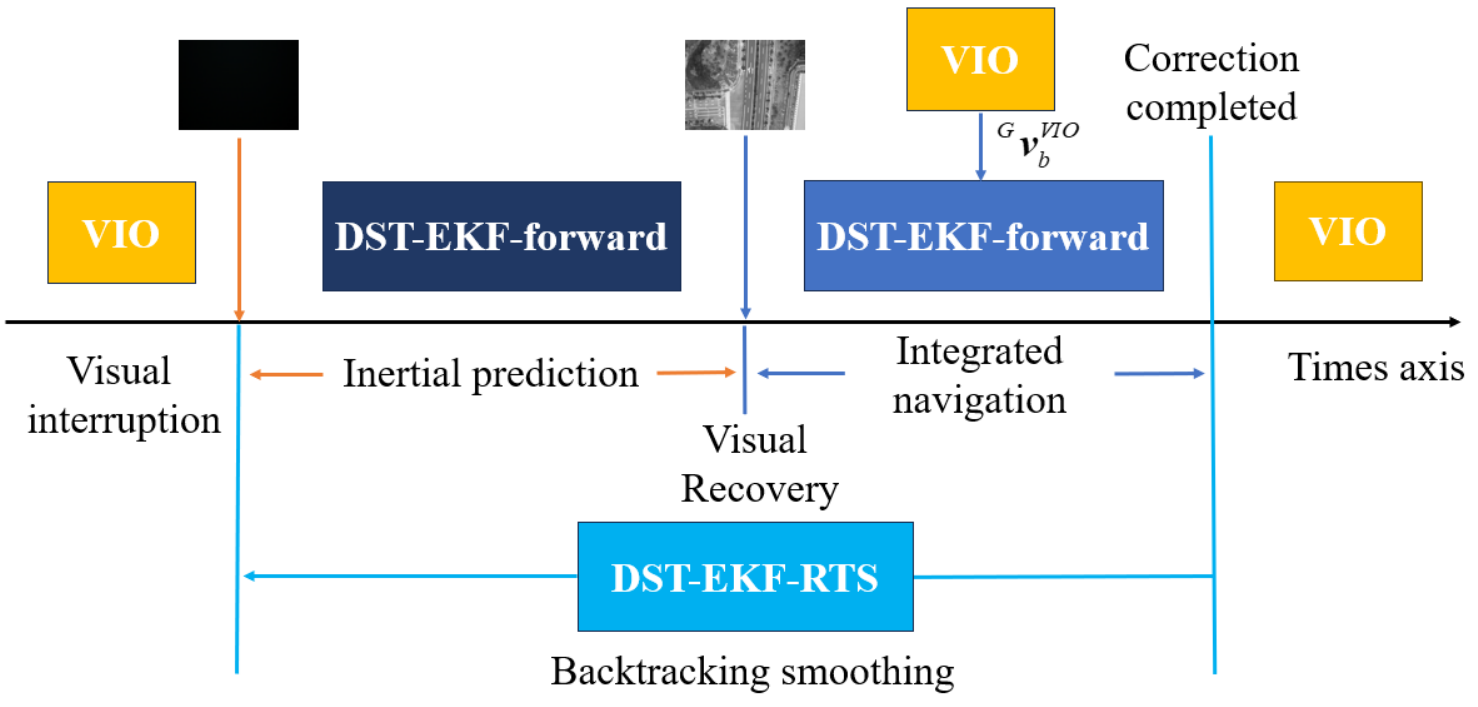} 
	\caption{
		Timing chart of RTS backtracking smoothing under discontinuous observations.
	}
	\label{fig:RTSintegram T}
\end{figure}

Considering that there is a scale deviation in the system's estimation after visual recovery, we add the scale error $\delta \boldsymbol{s}$ to the error state vector of \eqref{eqn:delta xb state}:
\begin{equation}
    \begin{aligned}
        \delta \boldsymbol{x}_{b}&=\begin{bmatrix} _{{b}_{k}}^{G}\boldsymbol{\phi}^{T} & \delta ^{G}\boldsymbol{v}_{{b}_{k}}^{T} & \delta ^{G}\boldsymbol{p}_{{b}_{k}}^{T} & \delta \boldsymbol{b}_{{g}_{k}}^{T} & \delta \boldsymbol{b}_{{a}_{k}}^{T} & {\delta \boldsymbol{s}}^{T} \end{bmatrix} ^{T}
    \end{aligned}
\end{equation}

Introducing the body frame's velocity after visual recovery, the measurement model of DST-EKF-forward can be defined as:
\begin{equation} \label{eqn:RTS Z}
    \begin{aligned}
        \boldsymbol{Z}_{v} &= {\boldsymbol{H}_{v}}{\delta \boldsymbol{x}_{v}} + \boldsymbol{n}_{v} \\
        \boldsymbol{Z}_{v}&={^{b}{\boldsymbol{\widetilde v}_{IMU}}} - {^{b}{\boldsymbol{\widetilde v}}_{VIO}} \\
	&= {^{b}{\boldsymbol{\widetilde v}_{IMU}}} - {^{b}{\boldsymbol{v}}_{VIO}} - {^{b}{\boldsymbol{\widetilde v}}_{VIO}}{\delta \boldsymbol{s}} \\ 
	&\approx {_{G}^{b}\boldsymbol{R}}{\delta^{G}\boldsymbol{v}_{{b}}} - {^{b}{\boldsymbol{\widetilde v}}_{VIO}}{\delta \boldsymbol{s}}   
    \end{aligned}
\end{equation}
where $\boldsymbol{n}_{v}$ is the measurement noise, ${^{b}{\boldsymbol{\widetilde v}_{IMU}}}$ is the velocity estimated by the INS solution, and ${^{b}{\boldsymbol{\widetilde v}}_{VIO}}$ is the velocity provided by the VIO system after visual recovery.

As shown in \eqref{eqn:RTS Z}, the measurement jacobian matrix $\boldsymbol{H}_{v}$ can be represented as:
\begin{align}\label{eqn:RTS H}
	\boldsymbol{H}_{v}=
	\begin{bmatrix}
		{\boldsymbol{0}_{3 \times 3}} & {_{G}^{b}\boldsymbol{R}} & {\boldsymbol{0}_{3 \times 9}} & - {^{b}{\boldsymbol{\widetilde v}}_{VIO}}
	\end{bmatrix} 
\end{align}

The algorithm flow of DST-EKF-forward is shown in \algref{alg:DST-EKF-forward}.
\begin{algorithm}[htbp]
	\caption{DST-EKF-forward}
	\label{alg:DST-EKF-forward}
	\begin{algorithmic}[1]
		\REQUIRE ${\boldsymbol{\hat x}_{k-1}}, {\boldsymbol{P}_{f,k-1}}, {\boldsymbol{F}(t)}, {\boldsymbol{G}(t)} $                  
		\ENSURE ${\boldsymbol{\hat x}_{f,k}}, {\boldsymbol{\hat x}_{f,k/k-1}}, {\boldsymbol{P}_{f,k}}, {\boldsymbol{P}_{f,k/k-1}}, {\boldsymbol{\Phi}_{k,k-1}} $       
		\FOR {$k=1,2...,N$}
		\STATE Discretization of continuous systems: \\ 
		${\boldsymbol{F}(t) \rightarrow \boldsymbol{\Phi}_{k,k-1}}$, ${\boldsymbol{G}(t) \rightarrow \boldsymbol{\Gamma}_{k-1}}$.            
		\STATE One-step state prediction: \\ 
		${{\boldsymbol{\hat x}_{f,k/k-1}} = {\boldsymbol{\Phi}_{k,k-1}}} {\boldsymbol{\hat x}_{k-1}} $
		\STATE The state one-step prediction covariance matrix: \\ 
		${\boldsymbol{P}_{f,k/k-1}} = {\boldsymbol{\Phi}_{k,k-1}} {\boldsymbol{P}_{f,k-1}} {\boldsymbol{\Phi}_{k,k-1}^{T}} + {\boldsymbol{\Gamma}_{k-1}} {\boldsymbol{Q}_{k-1}} {\boldsymbol{\Gamma}_{k-1}^{T}} $
		\IF {Visual recovery?}                        
		\STATE State estimation: ${\boldsymbol{\hat x}_{f,k}} = {\boldsymbol{\hat x}_{f,k,k-1}}$
		\STATE Covariance matrix: ${\boldsymbol{P}_{f,k}} = {\boldsymbol{P}_{f,k,k-1}}$  
		\ELSE
		\STATE Filtering gain: \\
		${\boldsymbol{K}_{f,k}} = {\boldsymbol{P}_{f,k/k-1}}{\boldsymbol{H}_{f,k}^{T}}{({\boldsymbol{H}_{f,k}}{\boldsymbol{P}_{f,k/k-1}}{\boldsymbol{H}_{f,k}^{T}}+{\boldsymbol{R}_{k}})^{-1}}$
		\STATE State estimation: \\
		${\boldsymbol{\hat x}_{f,k}} = {\boldsymbol{\hat x}_{f,k/k-1}} + {\boldsymbol{K}_{f,k}}({\boldsymbol{z}_{k}}-{\boldsymbol{H}_{f,k}} {\boldsymbol{\hat x}_{f,k,k-1}} ) $
		\STATE Covariance matrix: \\
		${\boldsymbol{P}_{f,k}} = {\boldsymbol{A}}{\boldsymbol{P}_{f,k/k-1}}{\boldsymbol{A}^{T}} + {{\boldsymbol{K}_{f,k}}}{\boldsymbol{R}_{k}}{{\boldsymbol{K}_{f,k}^{T}}} $ \\
		${{\boldsymbol{A}} = {\boldsymbol{I}}-{\boldsymbol{K}_{f,k}}{\boldsymbol{H}_{f,k}}}$
		\ENDIF
		\ENDFOR
	\end{algorithmic}
\end{algorithm}

After the forward filtering state estimation is stable, RTS method \cite{rts} is used for backtracking optimization estimation. The core formula is as follows \cite{rts}:
\begin{align}\label{eqn:RTS core}
	{\boldsymbol{K}_{RTS,k}} &= {\boldsymbol{P}_{f,k}}{\boldsymbol{\Phi}_{k+1/k}^{T}}{\boldsymbol{P}_{f,k+1/k}^{-1}} \\
	{\boldsymbol{\hat x}_{RTS,k}} &= {\boldsymbol{\hat x}_{f,k}} + {\boldsymbol{K}_{RTS,k}}({\boldsymbol{\hat x}_{RTS,k+1}} - {\boldsymbol{\hat x}_{f,k+1/k}}) \\
	{\boldsymbol{P}_{RTS,k}} &= {\boldsymbol{P}_{f,k}} + {\boldsymbol{K}_{RTS,k}}{({\boldsymbol{P}_{RTS,k+1}} - {\boldsymbol{P}_{f,k+1/k}})}{\boldsymbol{K}_{RTS,k}^{T}}
\end{align}
where $\{{\boldsymbol{\hat x}_{f}}, {\boldsymbol{P}_{f}}, {\boldsymbol{\Phi}} \}$ is the relevant parameters of DST-EKF-forward, and $\{{\boldsymbol{\hat x}_{RTS}}, {\boldsymbol{P}_{RTS}} \}$ is the relevant parameters of DST-EKF-RTS. The algorithm flow of DST-EKF-RTS is shown in \algref{alg:DST-EKF-RTS}.
\begin{algorithm}[htbp]
	\caption{DST-EKF-RTS}
	\label{alg:DST-EKF-RTS}
	\begin{algorithmic}[1]
		\REQUIRE ${\boldsymbol{\hat x}_{f,k}}, {\boldsymbol{\hat x}_{f,k+1/k}},{\boldsymbol{P}_{f,k}}, {\boldsymbol{P}_{f,k+1/k}}, {\boldsymbol{\Phi}_{k+1/k}} $                  
		\ENSURE ${\boldsymbol{\hat x}_{RTS,k}}, {\boldsymbol{P}_{RTS,k}} $       
		\STATE initialization: ${\boldsymbol{P}_{RTS,N} = \boldsymbol{P}_{f,N}}$, ${\boldsymbol{\hat x}_{RTS,N} = \boldsymbol{\hat x}_{f,N}}$ 
		\FOR {$k=N-1,N-2...1$}
		\STATE Filtering gain of RTS: \\
		${\boldsymbol{K}_{RTS,k}} = {\boldsymbol{P}_{f,k}}{\boldsymbol{\Phi}_{k+1/k}^{T}}{\boldsymbol{P}_{f,k+1/k}^{-1}} $
		\STATE State estimation of RTS: \\
		${\boldsymbol{\hat x}_{RTS,k}} = {\boldsymbol{\hat x}_{f,k}} + {\boldsymbol{K}_{RTS,k}}({\boldsymbol{\hat x}_{RTS,k+1}} - {\boldsymbol{\hat x}_{f,k+1/k}}) $
		\STATE Covariance matrix of RTS: \\
		${\boldsymbol{P}_{RTS,k}} = {\boldsymbol{P}_{f,k}} + {\boldsymbol{K}_{RTS,k}}{\boldsymbol{B}}{\boldsymbol{K}_{RTS,k}^{T}} $ \\
		${{\boldsymbol{B}} = {\boldsymbol{P}_{RTS,k+1}} - {\boldsymbol{P}_{f,k+1/k}}}$
		\ENDFOR		
	\end{algorithmic}
\end{algorithm}

In summary, combined with the enhanced DST-RTS smoothing strategy, our proposed SP-VIO can correct accumulated errors after visual recovery, and does not depend on the loop closure.


\section{Experimental Evaluation}\label{sec:Experiment Evaluation}
This section provides a comprehensive performance evaluation of our proposed SP-VIO, including the ablation experiments and comparisons with the current SOTA VIO algorithms, such as OpenVINS \cite{9196524} and VINS-Mono \cite{WOS:000442341000003}.
Real-world experiments include both open-source and personal datasets, and all contain time-synchronized IMU data, images, and ground truth, where a detailed overview is provided in \tabref{tab:Navigation Sensors}.
To quantitatively evaluate the localization accuracy of the relevant algorithms, we calculate absolute translation error (ATE) \cite{6385773} across the entire experimental trajectory using the EVO tools \cite{WOS:000458872706092}.
All experiments are conducted on a standard laptop (Intel Core i7-10875H CPU @ 2.80GHZ).
Since the RTS relocalization algorithm is mainly used for discontinuous observation conditions, we dedicate a subsection to evaluating it.

\begin{table*}[htbp]
	\centering
	\caption{Navigation Sensors Specifications}
	\label{tab:Navigation Sensors}
	\resizebox{1\textwidth}{!}{
		\begin{tabular}{ccccccc}
			\toprule
			Dataset & Environment & Carrier & IMU & Camera & Time synchronization & Ground truth\\
			\midrule
			EuRoC\cite{WOS:000382981300001} & indoors & MAV & \makecell{stereo gray \\ $752\times480$@$20$Hz} & \makecell{ADIS16488 6-axis \\ acc/gyro @200Hz} & hardware & \makecell{motion capture pose \\ @100Hz}\\
			Tum VI\cite{8593419} & in/outdoors & handheld & \makecell{stereo gray \\ $1024\times1024$@$20$Hz}  &  \makecell{BMI160 $6$-axis \\ acc/gyro @$200$Hz}  & hardware & \makecell{partial motion capture pose \\ @$120$Hz}\\
			Kitti Odometry\cite{WOS:000309166203066} & outdoors & car & \makecell{stereo gray \\ $1024\times768$@$10$Hz} & \makecell{OXTS RT 3003 6-axis \\ acc/gyro @100Hz} & software & \makecell{OXTS RT $3003$ pose \\ @$10$Hz}\\
			\midrule
			\multirow{2}{*}{Nudt VI} & \multirow{2}{*}{outdoors} & car & \makecell{mono gray \\ $1620\times1220$@$20$Hz} & \makecell{STIM300 6-axis \\ acc/gyro @200Hz} & hardware & \makecell{fused RTK/IMU pose \\ @10Hz}\\
			& & UAV & \makecell{mono gray \\ $720\times540$@$10$Hz} & \makecell{EIMU910 6-axis \\ acc/gyro @200Hz} & hardware & \makecell{fused RTK/IMU pose \\ @10Hz}\\
			\bottomrule
		\end{tabular}
	}
	
\end{table*}

\subsection{Ablation experiments}\label{sec:Ablation experiments on SP-VIO}
To verify our proposed algorithm,  ablation experiments were first performed on the simulation trajectory and KITTI Odometry dataset \cite{WOS:000309166203066}, and the comparison algorithms included: Std-EKF(baseline system)\cite{9196524}, DST-EKF(modified by \secref{sec:System model}), and PO-EKF(modified by \secref{sec:Measurement Model Based on PO Visual description}).

\subsubsection{Monte-Carlo Simulations}\label{sec:Monte-Carlo Simulations}
\begin{figure*}[htbp]
	\centering
	\subfigure[SIM\_UAV] { \centering \includegraphics [width=2.3in] {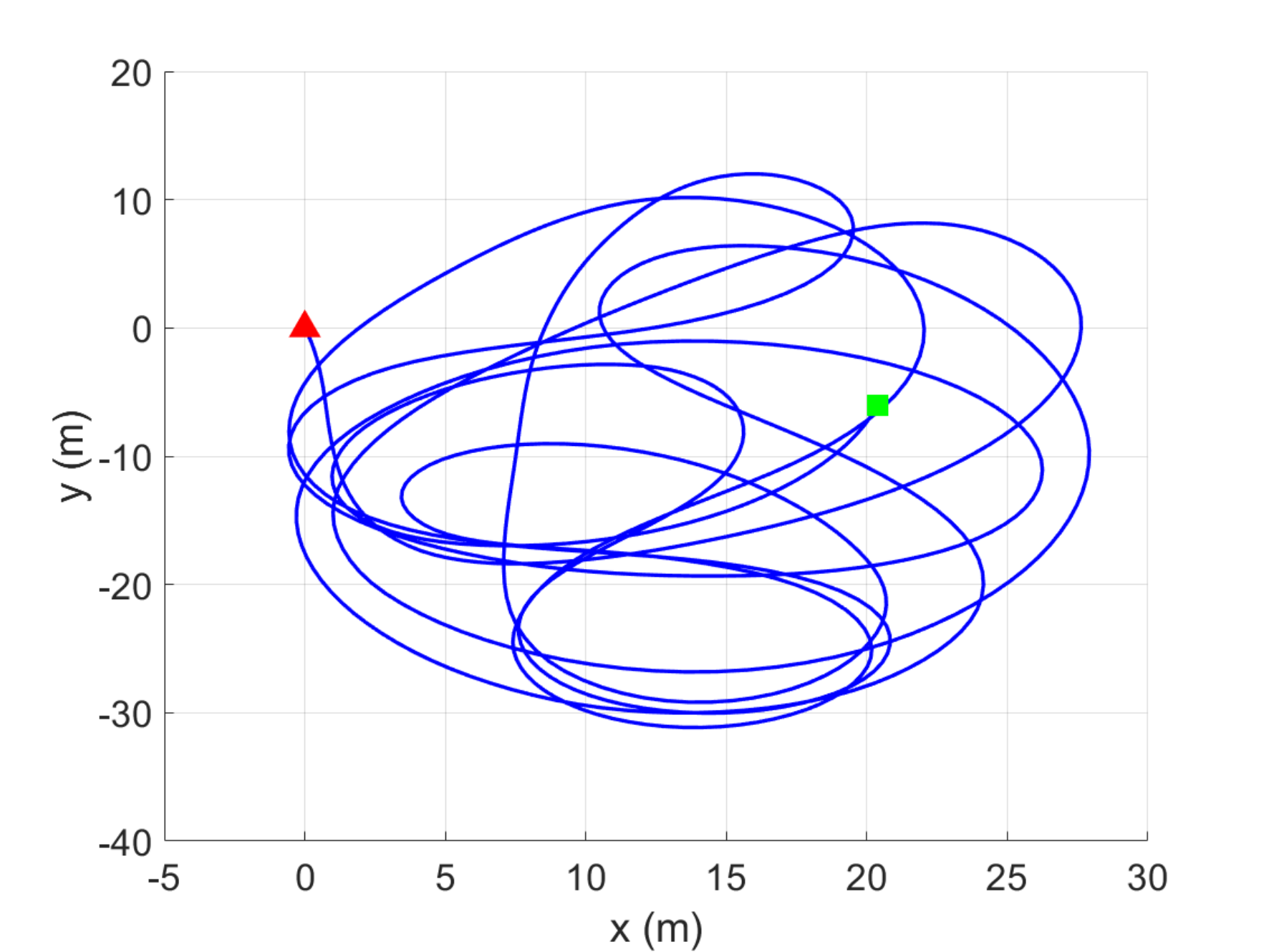}\label{fig:SIM_UAV}} 
	\subfigure[SIM\_Pedestrian] { \centering \includegraphics [width=2.3in] {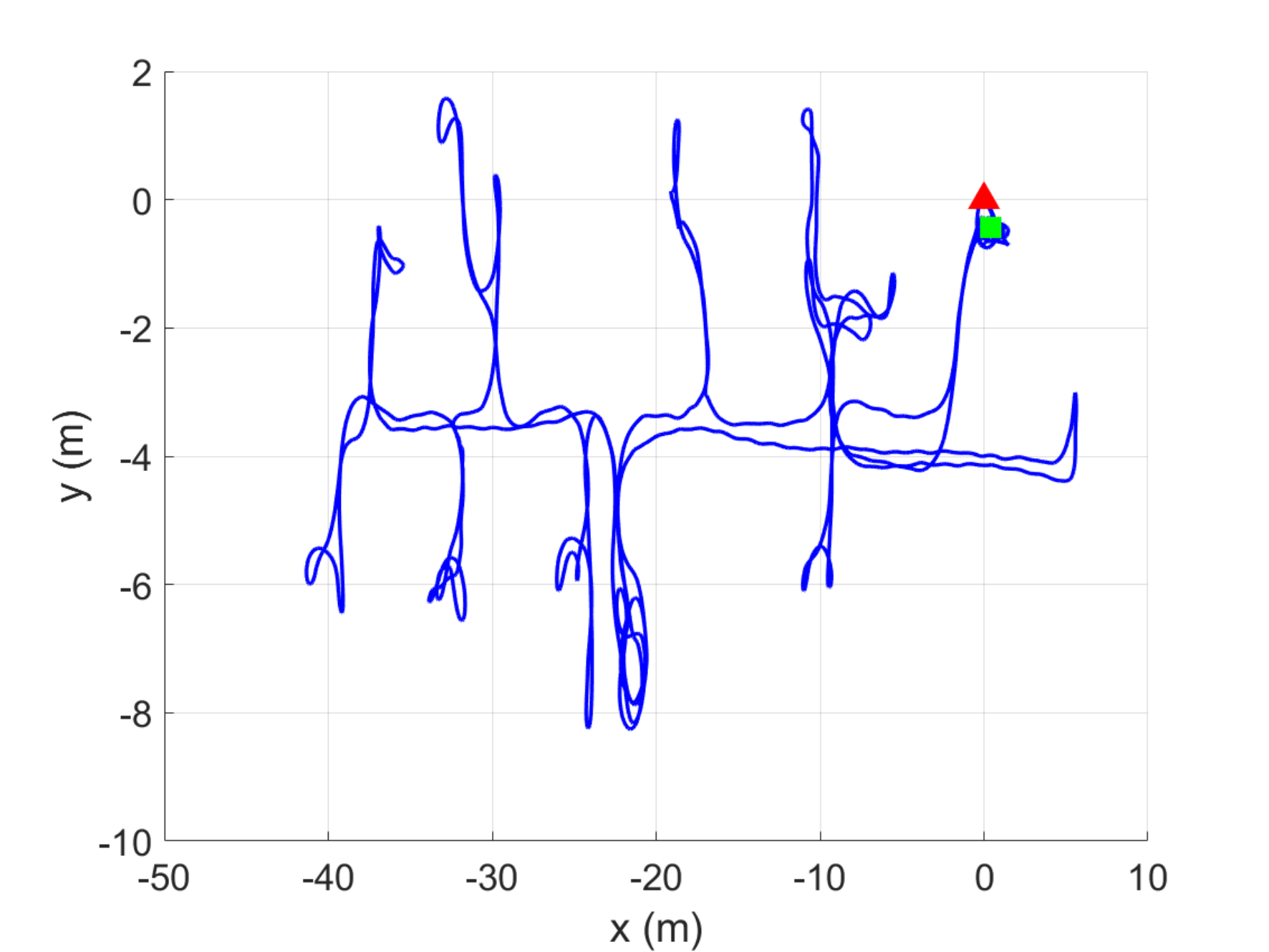}\label{fig:SIM_pedestrian}}
	\subfigure[SIM\_CAR] { \centering \includegraphics [width=2.3in] {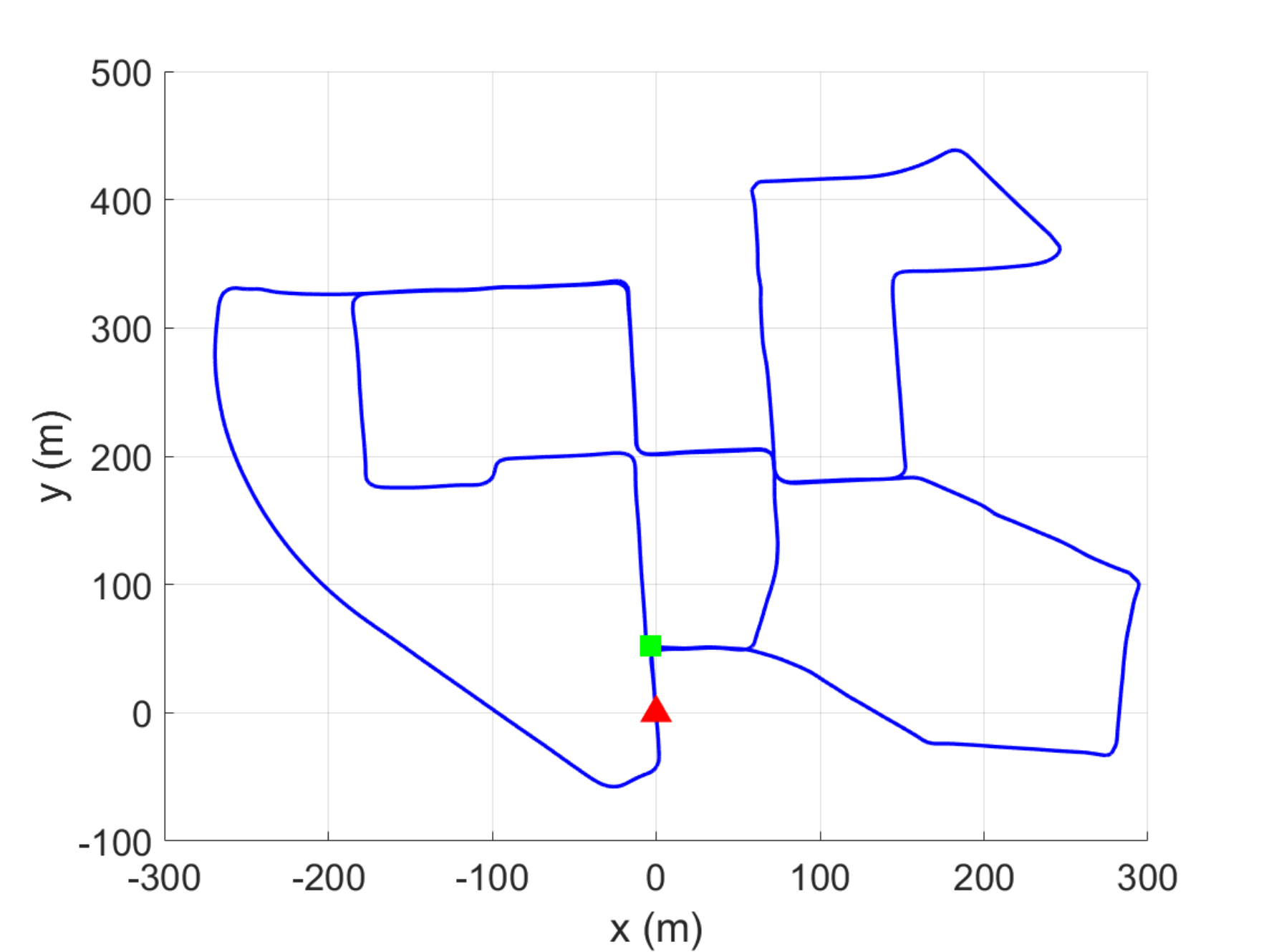}\label{fig:SIM_CAR}}
	\caption{
		Simulated trajectories. Left: SIM\_UAV, middle: SIM\_Pedestrian and right: SIM\_CAR . Note that the red triangular denotes the start of the trajectory while the green square indicates the end.      
	}
	\label{fig:simulated trajectories}
\end{figure*}

The simulation trajectory is shown in \figref{fig:simulated trajectories}, corresponding to the three carriers of UAV, handheld device and commercial vehicle respectively, which can cover most motion situations in the real world environment. 
The inertial and camera measurements of the simulation are generated by \cite{9196524}, and the basic configuration is shown in \tabref{tab:Monte-Carlo Simulation Parameters}.

\begin{figure*}[htbp]
	\centering
	\subfigure[SIM\_UAV] { \centering \includegraphics [width=2.3in] {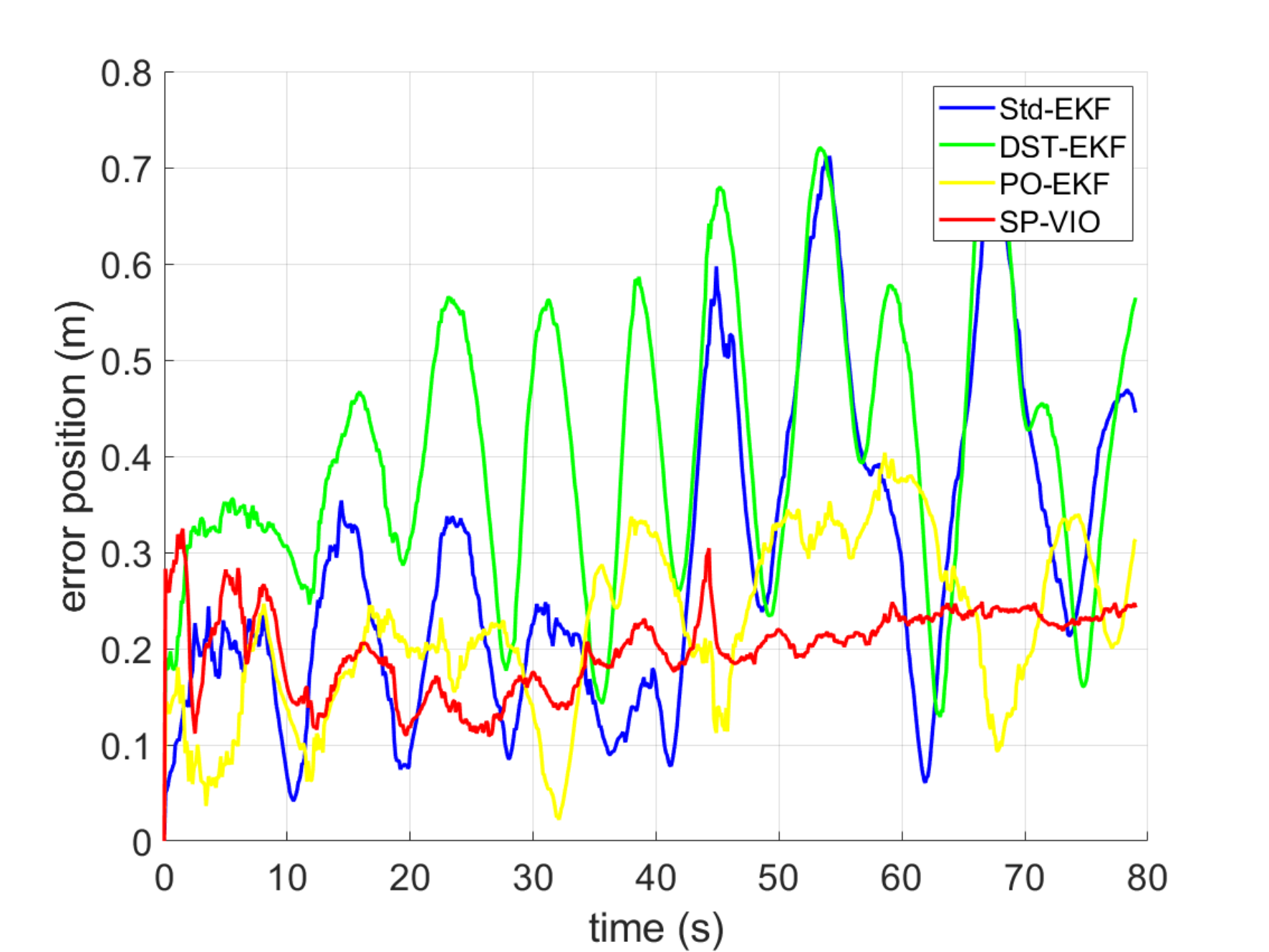}\label{fig:SIM_UAV_rm}} 
	\subfigure[SIM\_Pedestrian] { \centering \includegraphics [width=2.3in] {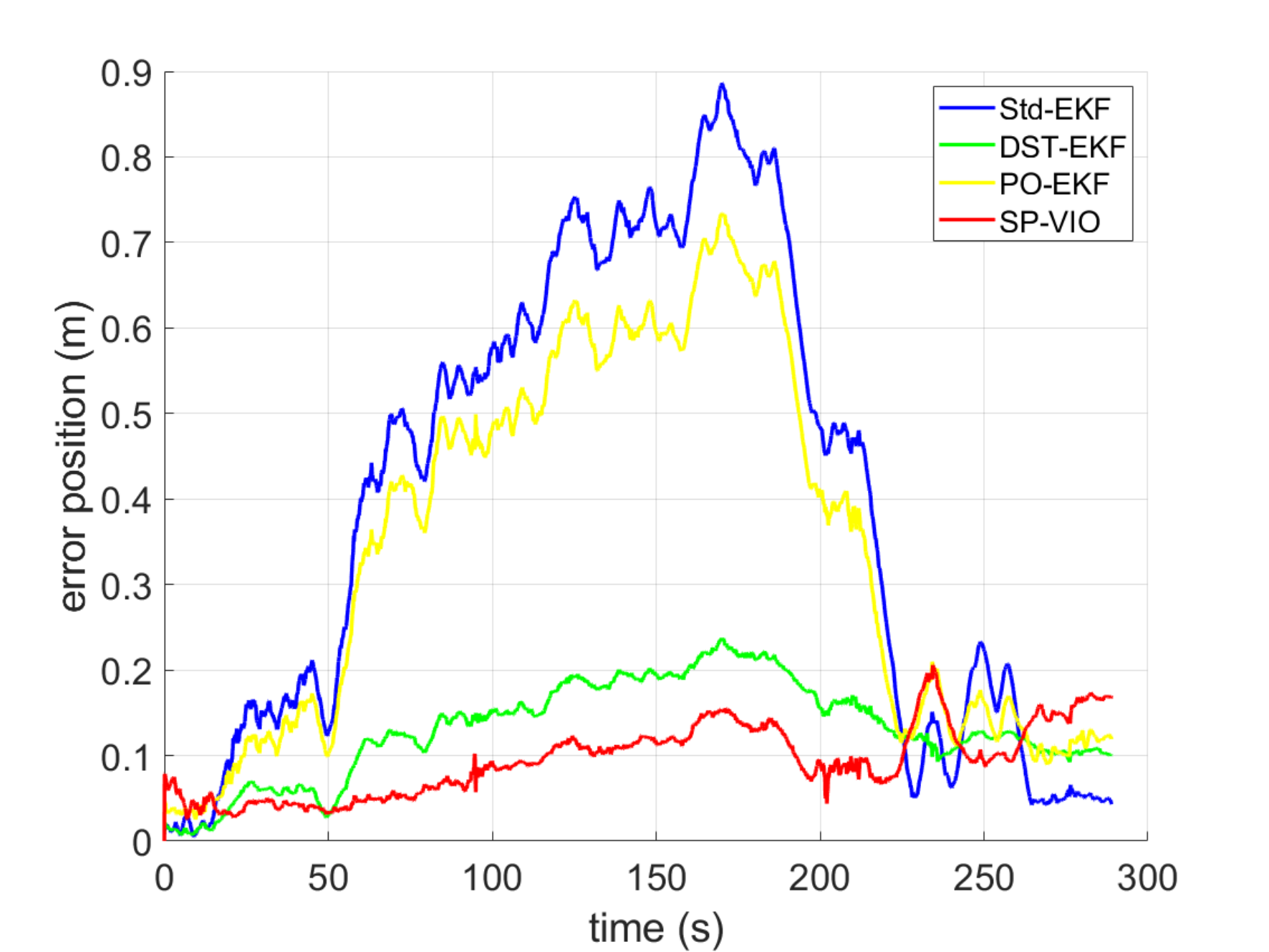}\label{fig:SIM_pedestrian_rm}}
	\subfigure[SIM\_CAR] { \centering \includegraphics [width=2.3in] {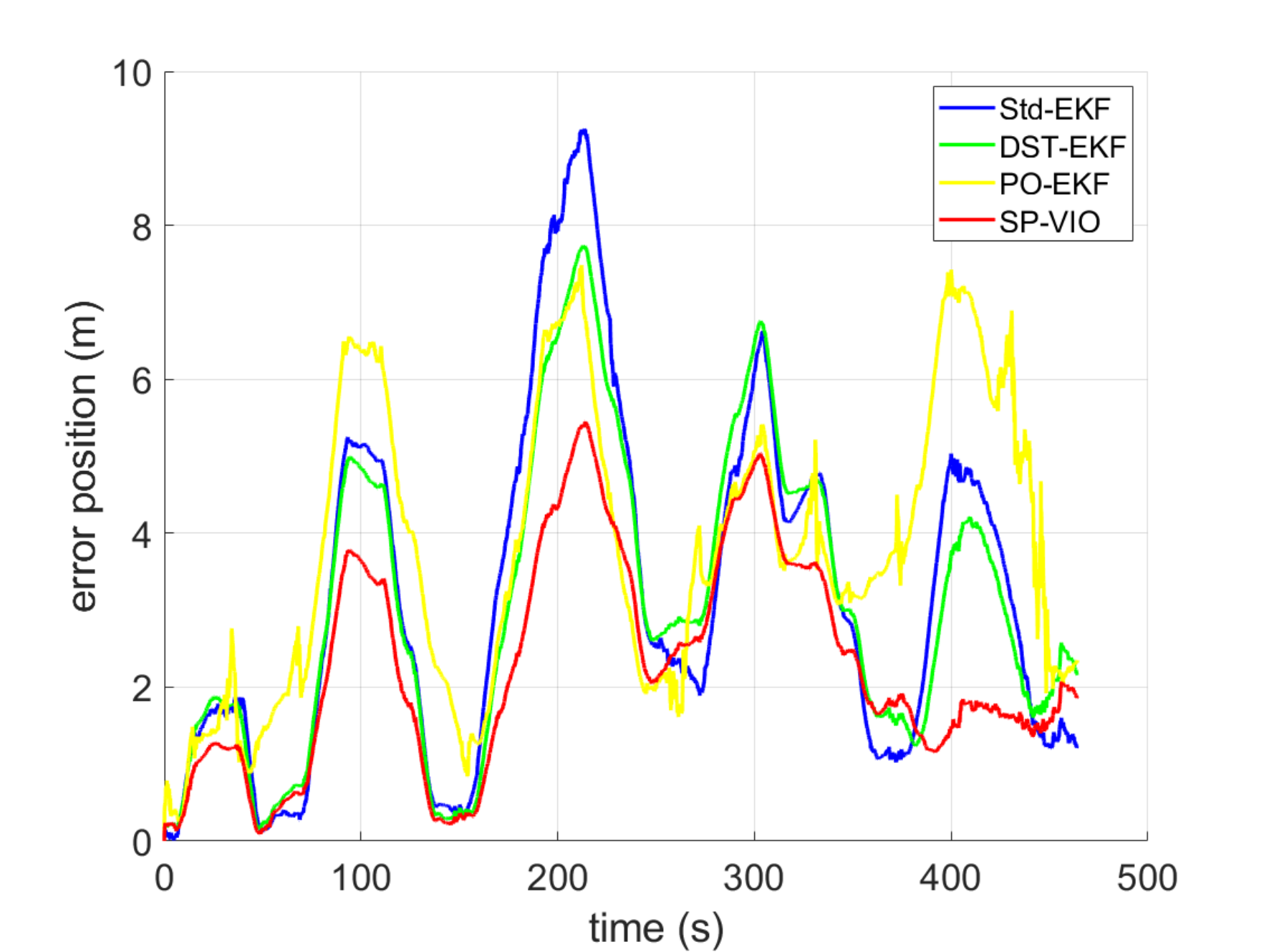}\label{fig:SIM_CAR_rm}}
	\caption{
		Average localization error for 50 Monte-Carlo runs of algorithms(Std-EKF, PO-EKF, DST-EKF, and SP-VIO) on different simulation trajectories, respectively.      
	}
	\label{fig:RMSE SIM}
\end{figure*}

The overall results are shown in \tabref{tab:Simulated ATE} and \figref{fig:RMSE SIM}.
From the results, SP-VIO showed the best localization performance on all simulation trajectory thanks to the reconstructed system model and measurement model, which allowed it to output more consistent navigation results under different motion states. 
In addition, DST-EKF and PO-EKF show suboptimal performance, but are not satisfactory on some sequences, which is consistent with the observability analysis results in \secref{sec:Observability Analysis}.

\begin{table}[htbp]
	\begin{center}
		\caption{Monte-Carlo Simulation Parameters}
		\label{tab:Monte-Carlo Simulation Parameters}
		\begin{threeparttable}
				\begin{tabular}{cccc}
					\toprule
					Parameter & Value & Parameter & Value  \\			
					\midrule
					IMU Freq. (hz) & $400$ & Cam Freq. (hz) & $10$ \\
					Gyro White Noise & $1.6968e^{-4}$ & Gyro Random Walk & $1.9393e^{-4}$ \\
					Accel. White Noise & $2.0000e^{-3}$ & Accel. White Noise & $2.0000e^{-3}$ \\
					Max Cam Pts/Frame & $100$ & Max Cam Pts/Frame & $40$ \\
                        Max Clone Size & $11$ & Pt Feat. Rep. & GLOBAL \\
                        Cam noise (pixel) & $1$ & Cam noise (pixel) & false \\
					\bottomrule
				\end{tabular}
		\end{threeparttable}	
	\end{center}
\end{table}

\begin{table}[htbp]
	\begin{center}
		\caption{RMSE for 50 Monte-Carlo Runs on 3 Simulated Trajectories in Meters}
		\label{tab:Simulated ATE}
		\begin{threeparttable}
					\resizebox{0.5\textwidth}{!}{
				\begin{tabular}{cccccc}
					\toprule
					Sequence & Distance & Std-EKF & DST-EKF & PO-EKF & SP-VIO  \\			
					\midrule
					SIM\_UAV & $662$ & $0.33$ & $0.43$ & $0.24$ & $\boldsymbol{0.20}$ \\
					SIM\_Pedestrian & $293$ & $0.48$ & $0.14$ & $0.41$ & $\boldsymbol{0.10}$ \\
					SIM\_CAR & $3678$ & $3.93$ & $3.64$ & $4.20$ & $\boldsymbol{2.65}$ \\
					Average & $1544$ & $1.58$ & $1.41$ & $1.62$ & $\boldsymbol{0.99}$ \\
					\bottomrule
				\end{tabular}
						}
		\end{threeparttable}	
	\end{center}
\end{table}

\subsubsection{Real-World Experiments}\label{sec:Real-World Experiments}
We further evaluate the SP-VIO with real-world KITTI benchmark, which includes different acquisition scenarios, such as urban, rural, and highway, with complex motion conditions and long distances.
In order to ensure the fairness of this experiment, we initialize the four algorithms by assigning values, and provide the same preprocessing features to avoid the influence of other variables on the experimental results.
Additionally, the imu data of '00' was lost in some periods, and '03' has been removed from the official website, so we chose another 9 sequences for the evaluation.

\begin{table}[htbp]
	\begin{center}
		\caption{RMSE in Kitti Odometry Dataset in Meters}
		\label{tab:Kitti RMSE ablation}
		\begin{threeparttable}
				\begin{tabular}{cccccc}
					\toprule
					Sequence & Distance &  Std-EKF & DST-EKF & PO-EKF & SP-VIO  \\			
					\midrule
					01 & $2457.57$ & $27.54$ & $20.06$ & $20.48$ & $\boldsymbol{19.27}$ \\
					02 & $2685.71$ & $14.35$ & $12.79$ & $13.01$ & $\boldsymbol{12.01}$ \\
					04 & $393.31$ & $0.64$ & $\boldsymbol{0.26}$ & $0.5$ & $\boldsymbol{0.26}$ \\
					05 & $2208.69$ & $16.23$ & $16.19$ & $15.95$ & $\boldsymbol{15.57}$ \\
					06 & $1219.59$ & $5.94$ & $2.78$ & $5.72$ & $\boldsymbol{2.7}$ \\
					07 & $692.83$ & $3.56$ & $3.02$ & $2.82$ & $\boldsymbol{2.8}$ \\
					08 & $3226.71$ & $19.34$ & $9.17$ & $13.97$ & $\boldsymbol{6.77}$ \\
					09 & $1710.34$ & $13.08$ & $7.55$ & $9.38$ & $\boldsymbol{6.67}$ \\
					10 & $921.22$ & $8.10$ & $6.32$ & $6.16$ & $\boldsymbol{6.05}$ \\
					Average & $1724$ & $12.09$ & $8.68$ & $9.77$ & $\boldsymbol{8.01}$ \\
					\bottomrule
				\end{tabular}
		\end{threeparttable}	
	\end{center}
\end{table}

In \tabref{tab:Kitti RMSE ablation}, we show in detail the trajectory errors of the four algorithms on different sequences and the percentage accuracy improvement of the modified three algorithms relative to MSCKF.
It can be found that the localization accuracy is improved by 25.84$\%$ and 18.31$\%$ respectively after using the new observation model and the system model, indicating that the two methods have played a role in improving the algorithm model. When two new models are used at the same time, the performance improvement is the best, reaching 33.75$\%$, which verifies the feasibility of applying the two new models at the same time and the advanced nature of SP-VIO.

\subsection{Performance Evaluation with SOTA Algorithms}\label{sec:Performance Evaluation with SOTA Algorithms}
Through ablation experiments, we demonstrated the performance improvement of SPVIO compared to the baseline system. So in this section, we conducted a comparison experiment with OpenVINS \cite{9196524} and VINS-Mono \cite{WOS:000442341000003}, which are the current SOTA filter-based and optimization-based VIO systems respectively.
Test datasets include both public and personal datasets, collected from three different carriers.
\subsubsection{Public datasets}\label{sec:Public datasets evaluation}
To make the experiment more convincing, the public datasets adopt EuRoC \cite{WOS:000382981300001} and TUM-VI\cite{8593419}, in which both SOTA algorithms have relevant test results on official documentation.
Notably, since only the room sequence provides the ground truth of the full trajectory in Tum-VI, so it is chosen for evaluation.

As shown in \figref{fig:Overview2}, SP-VIO demonstrated better localization performance, significantly outperforming the comparative SOTA algorithms on both EuRoC and TUM-VI.
Meanwhile, the calculation efficiency evaluation in \figref{fig:Overview3} and \tabref{tab:Runtime Evaluation on EuRoC} show that the running time of SP-VIO is roughly the same as that of OpenVINS, but much shorter than that of VINS-Mono, which maintains good real-time performance.

For a more intuitive comparison of localization accuracy, we show partial experimental results such as 'MH\_04\_difficult', 'V1\_03\_difficult', and 'Room3\_512\_16' in \figref{fig:Partial result trajectories in public datasets2}, where the result trajectory is aligned with the ground truth by evo\cite{WOS:000458872706092}.
Compared with OpenVINS and VINS-Mono, the trajectory produced by SP-VIO demonstrates a closer endpoint and better consistency with the ground truth.

\subsubsection{Personal dataset}\label{sec:Personal dataset evaluation}
\begin{figure}[htbp]
	\centering
	\subfigure[] { \begin{minipage}{4cm} \centering \includegraphics [height=3.2cm]{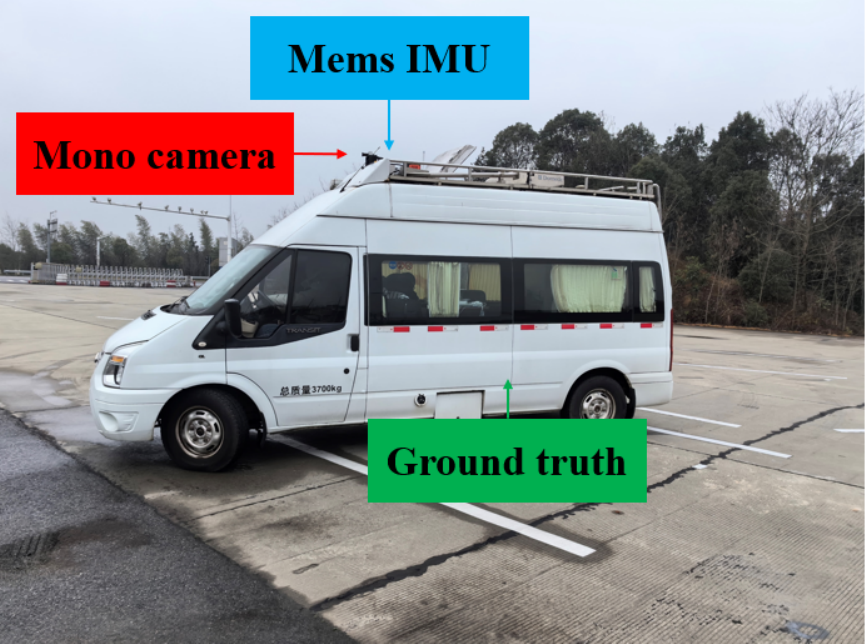}\label{fig:carplat} \end{minipage} } 
	\subfigure[] { \begin{minipage}{4cm} \centering \includegraphics [height=3.2cm]{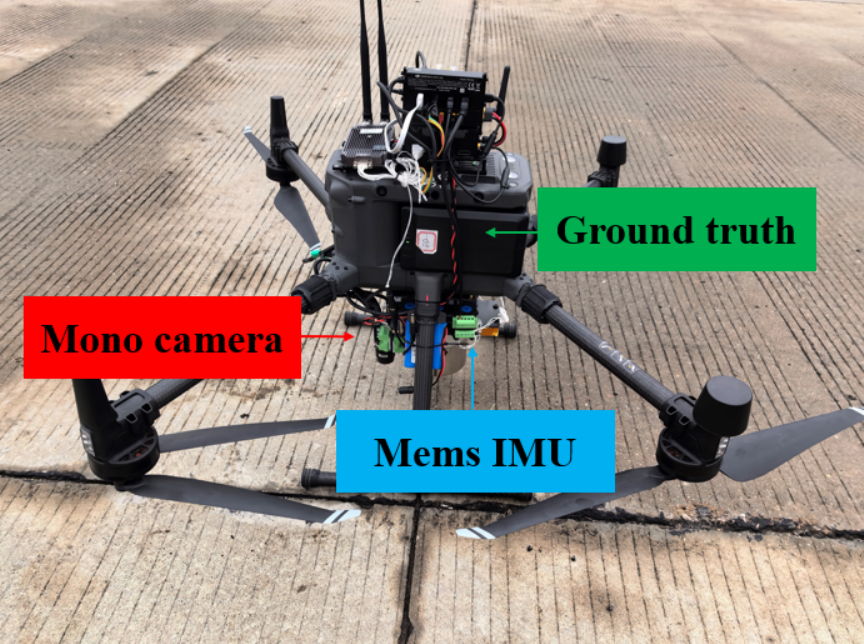}\label{fig:uavplat} \end{minipage} }
	\caption{
		Experimental platform and the sensor installation relationship in Nudt VI dataset.
		(a) Landvehicle experiment.
		(b) UAV experiment.
	}
	\label{fig:experimental platform}
\end{figure}

\begin{figure}[htbp]
	\centering
	\subfigure[] { \begin{minipage}{4cm} \centering \includegraphics [height=3.8cm]{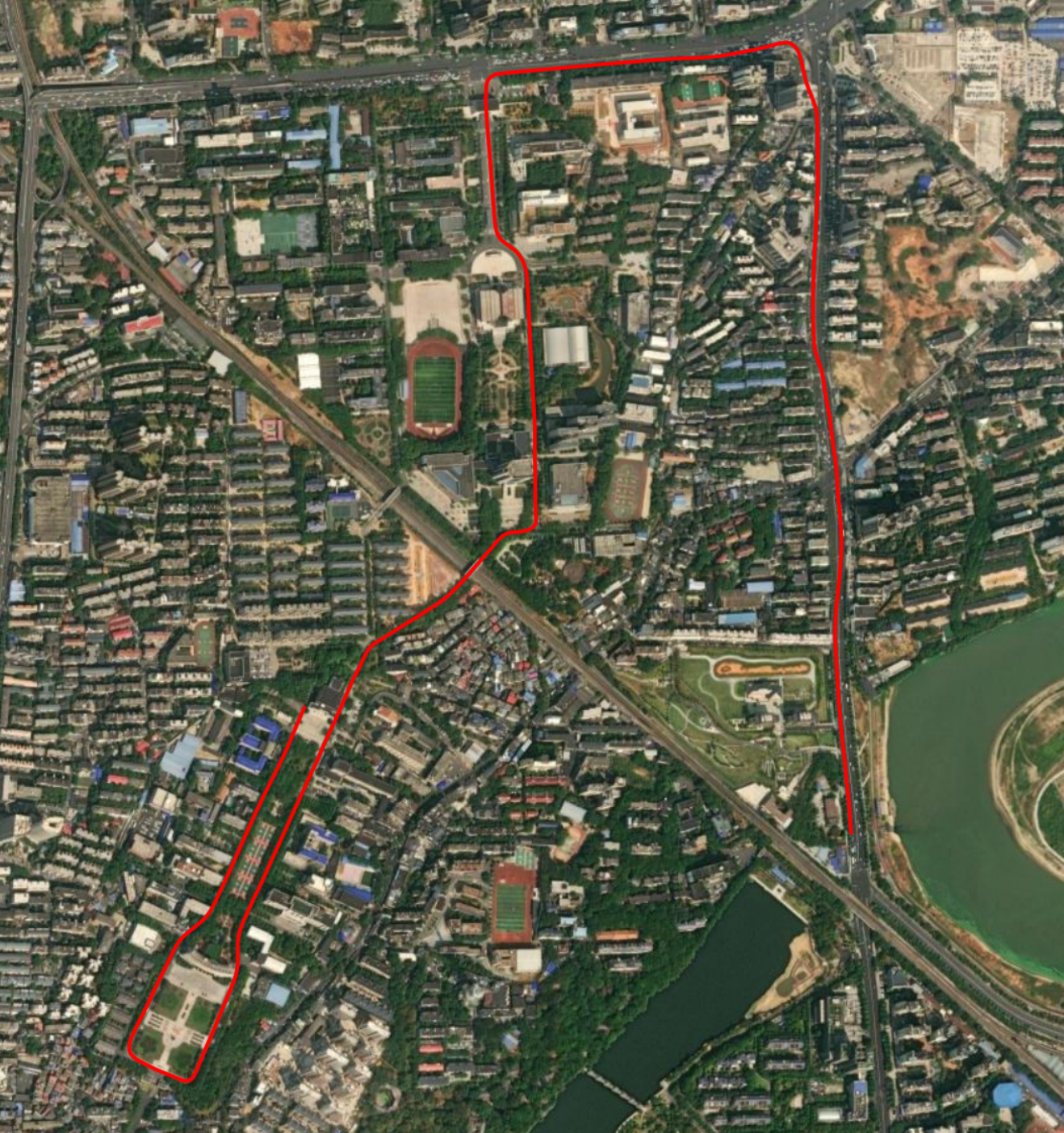}\label{fig:carworld} \end{minipage} } 
	\subfigure[] { \begin{minipage}{4cm} \centering \includegraphics [height=3.8cm]{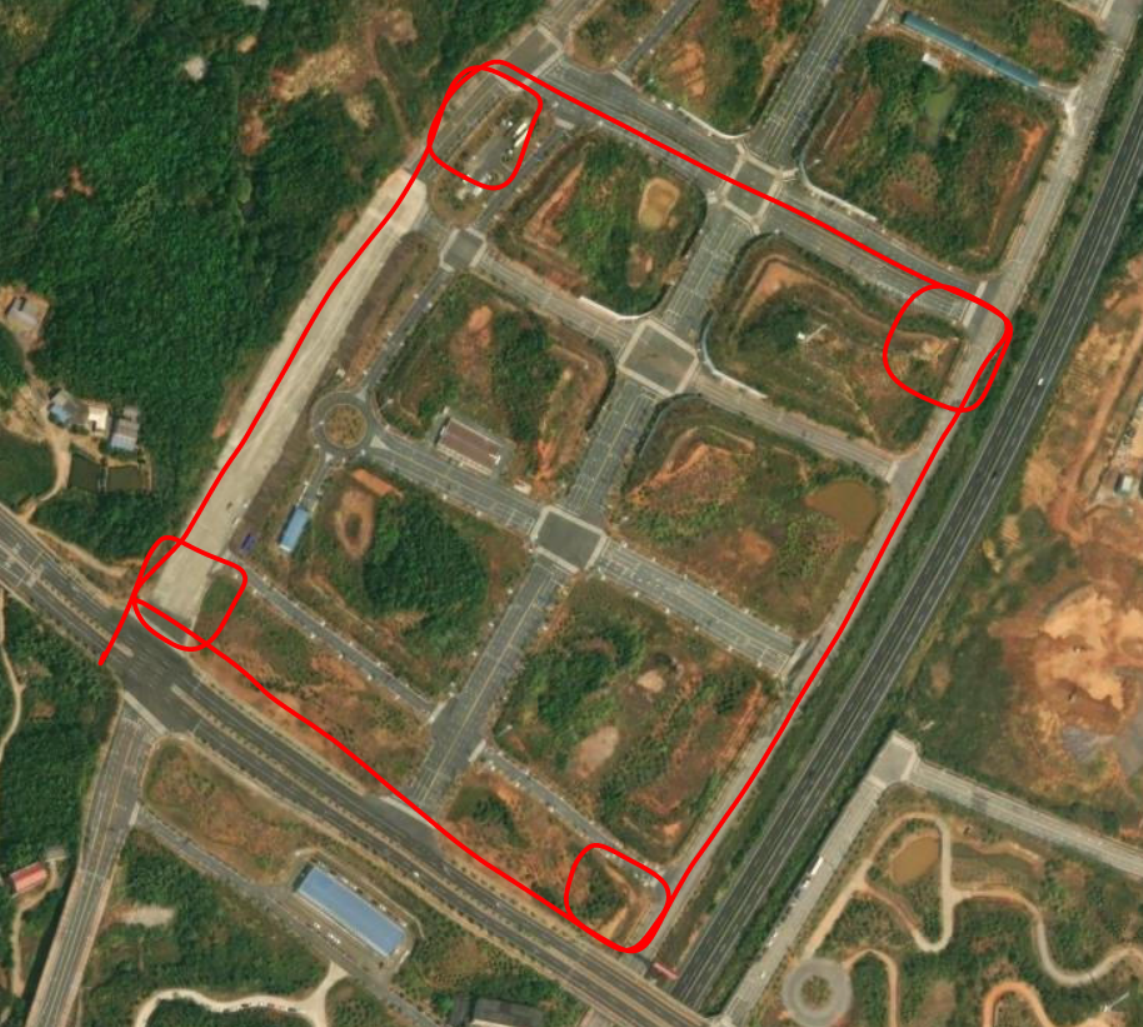}\label{fig:uavworld} \end{minipage} }
	\caption{
		Experimental trajectory in Nudt VI dataset.
		(a) Landvehicle experiment.
		(b) UAV experiment.
	}
	\label{fig:experimental wordld trajectory}
\end{figure}

\begin{figure}[htbp]
	\centering
	\subfigure[strong reflection] { \begin{minipage}{4cm} \centering \includegraphics [width=4cm,height=3cm]{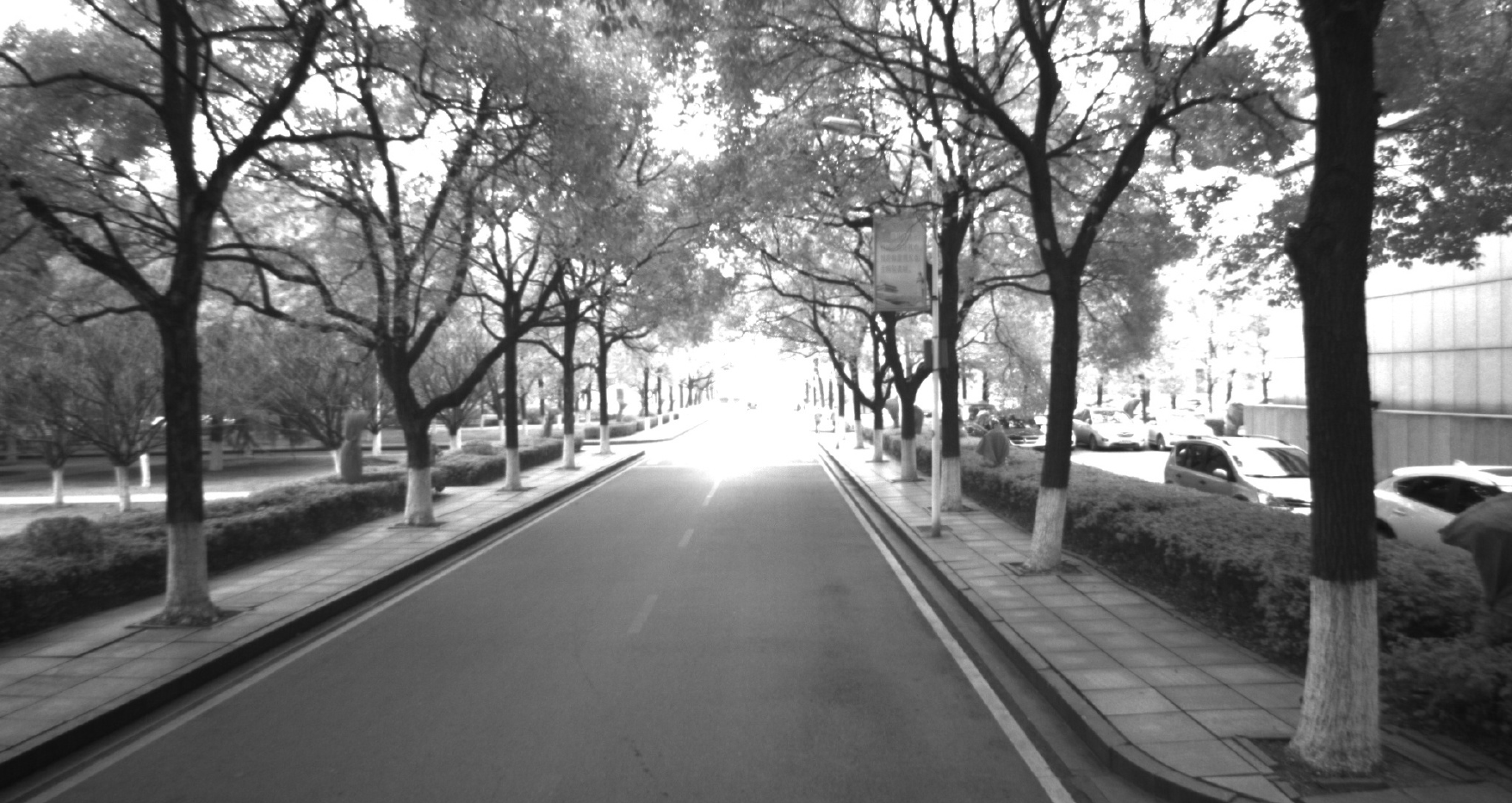}\label{fig:strong reflection} \end{minipage} } 
	\subfigure[complex traffic] { \begin{minipage}{4cm} \centering \includegraphics [width=4cm,height=3cm]{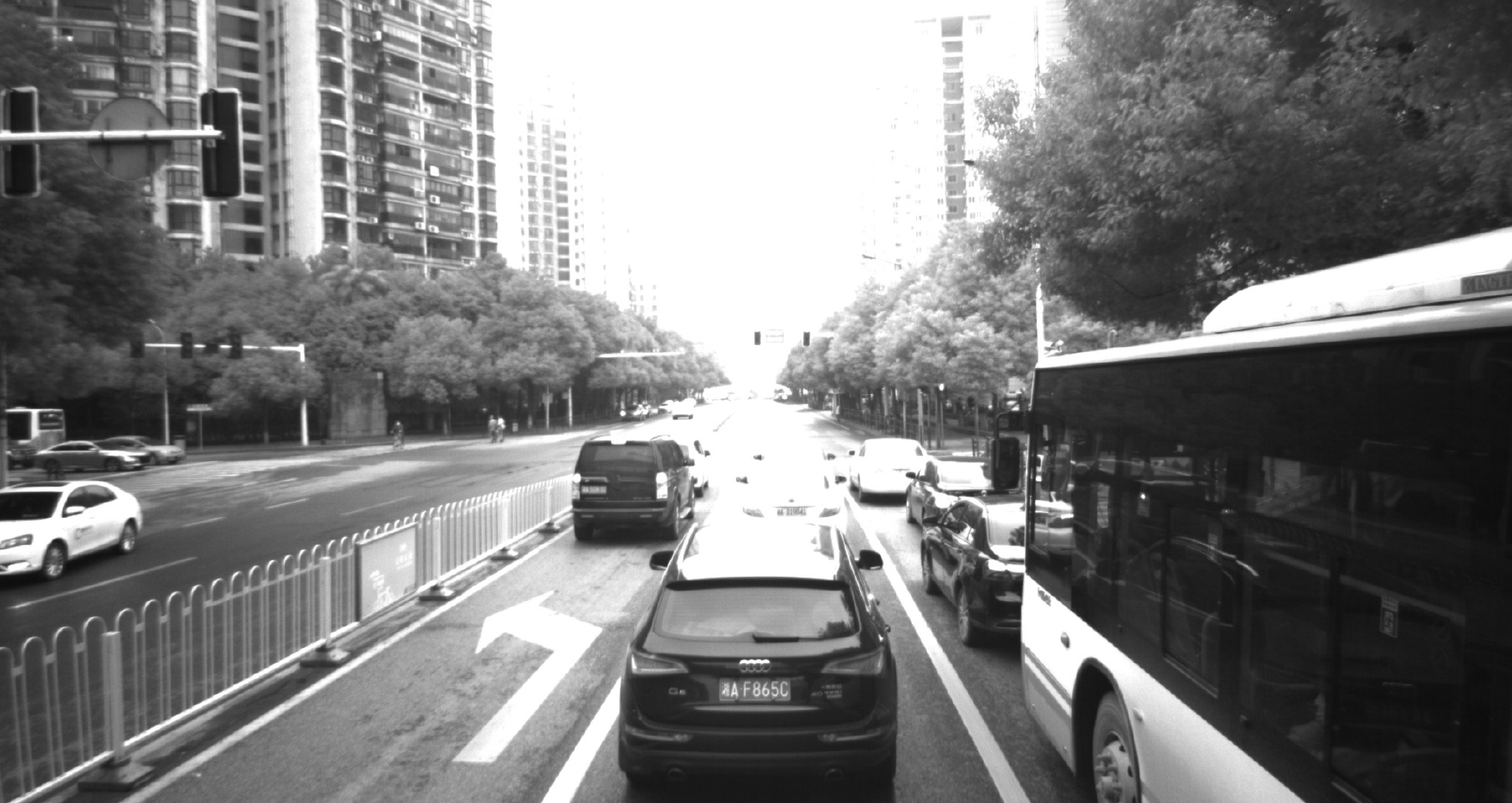}\label{fig:complex traffic} \end{minipage} } 
	\subfigure[low texture] { \begin{minipage}{4cm} \centering \includegraphics [width=4cm,height=3cm]{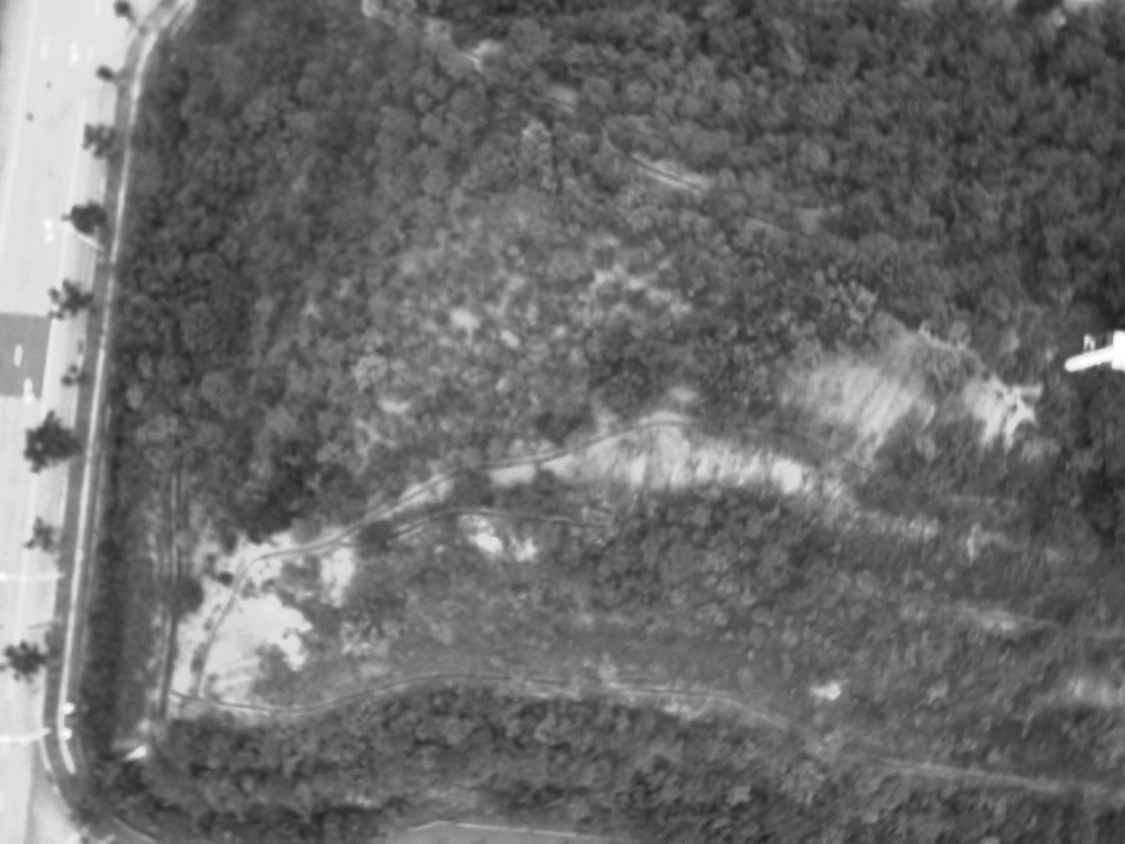}\label{fig:low texture} \end{minipage} }
	\subfigure[large depth] { \begin{minipage}{4cm} \centering \includegraphics [width=4cm,height=3cm]{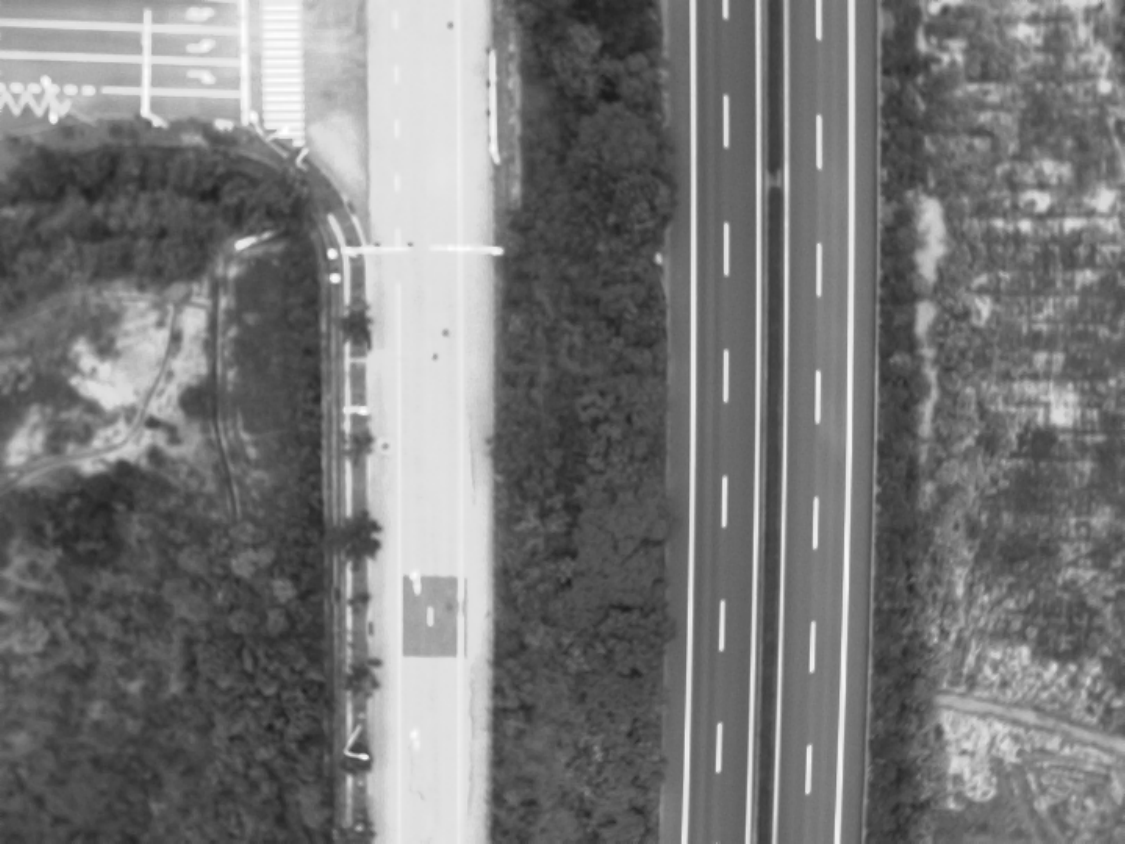}\label{fig:large depth} \end{minipage} }
	\caption{
		Challenging scenarios in Nudt VI dataset.
	}
	\label{fig:Challenging scenarios}
\end{figure}

The personal dataset Nudt-VI is collected from the commercial car and quadcopter UAV, and the sensor installation relationship and driving trajectory are shown in \figref{fig:experimental platform} and \figref{fig:experimental wordld trajectory} respectively.
This dataset contains scenes with strong reflection, complex traffic, low texture, large depth, etc., as shown in \figref{fig:Challenging scenarios}, which is challenging for VIO operation.

As an illustrative example, \figref{fig:comparison in nudt-vi car} depicts a comparison between the SP-VIO and SOTA algorithms for estimating the trajectory on 'nudt\_car'. It can be found that SP-VIO is closer to the ground truth, and the percentage error is only $0.29\%$ in the long distance experiment of 4.4km.

Based on the above experimental results, the root-mean-square errors (RMSE) of all sequences is listed in detail at \tabref{tab:Euroc RMSE}, which are evaluated by an ATE \cite{WOS:000458872706092,6385773}.

In summary, SP-VIO demonstrates superior localization performance compared to the other two SOTA algorithms on different experimental datasets.  This superiority is evident in both quantitative pose error metrics and qualitative trajectory smoothness assessments. In addition, SP-VIO also retains the real-time advantage of filter-based VIO and has high computational efficiency.

\begin{table}[htbp]
	\begin{center}
		\setlength{\tabcolsep}{5pt}
		\caption{RMSE in experimental Datasets in Meters}
		\label{tab:Euroc RMSE}
		\begin{threeparttable}
		\begin{tabular}{*{5}{c}}
			\toprule
			Sequence & Distance & VINS-Mono$^{1}$ & OpenVINS$^{2}$ & SP-VIO \\			
			\midrule
			MH\_01\_easy & $79.87$ & $0.15$ & $0.09$ & $\boldsymbol{0.07}$ \\
			MH\_02\_easy & $72.96$ & $0.15$ & $0.15$ & $\boldsymbol{0.13}$ \\
			MH\_03\_medium & $130.13$ & $0.22$ & $0.15$ & $\boldsymbol{0.13}$ \\
			MH\_04\_difficult & $91.75$ & $0.32$ & $0.23$ & $\boldsymbol{0.17}$ \\
			MH\_05\_difficult & $97.59$ & $\boldsymbol{0.30}$ & $0.46$ & $0.36$ \\
			V1\_01\_easy & $58.59$ & $0.08$ & $0.06$ & $\boldsymbol{0.04}$ \\
			V1\_02\_medium & $75.89$ & $0.11$ & $0.10$ & $\boldsymbol{0.09}$ \\
			V1\_03\_difficult & $78.98$ & $0.18$ & $0.07$ & $\boldsymbol{0.04}$ \\
			V2\_01\_easy & $36.5$ & $0.08$ & $0.07$ & $\boldsymbol{0.06}$ \\
			V2\_02\_medium & $83.22$ & $0.16$ & $0.07$ & $\boldsymbol{0.06}$ \\
			V2\_03\_difficult & $86.13$ & $0.27$ & $0.16$ & $\boldsymbol{0.15}$ \\
			\midrule
			Room1\_512\_16 & $146.79$ & $0.07$ & $0.06$ & $\boldsymbol{0.05}$ \\
			Room2\_512\_16 & $141.6$ & $\boldsymbol{0.07}$ & $0.10$ & $0.08$ \\
			Room3\_512\_16 & $135.52$ & $0.11$ & $0.09$ & $\boldsymbol{0.07}$ \\
			Room4\_512\_16 & $68.70$ & $0.04$ & $0.04$ & $\boldsymbol{0.03}$ \\
			Room5\_512\_16 & $131.64$ & $0.2$ & $0.08$ & $\boldsymbol{0.07}$ \\
			Room6\_512\_16 & $67.27$ & $0.08$ & $0.05$ & $\boldsymbol{0.04}$ \\
			\midrule
			Nudt\_car & $4483.54$ & $72.53$ & $49.97$ & $\boldsymbol{12.79}$ \\
			Nudt\_UAV & $2708.48$ & $27.77$ & $--$ & $\boldsymbol{17.22}$ \\
			\bottomrule
		\end{tabular}
		\begin{tablenotes}
			\footnotesize
			\item[1] The experimental results of VINS-Mono on the public datasets are from \cite{WOS:000442341000003} and \cite{8593419}, respectively .
			\item[2] The experimental results of Open-VINS on the public datasets are run in the configuration recommended by the official documentation \cite{9196524}.
		\end{tablenotes}
		\end{threeparttable}
	\end{center}
\end{table}

\begin{table*}[htbp]
	\centering
	\caption{Runtime Evaluation on Public Datasets}
	\label{tab:Runtime Evaluation on EuRoC}
	\resizebox{1\textwidth}{!}{
		\begin{tabular}{cccccccccc}
			\toprule
			\multirow{2}{*}{Sequence} & \multicolumn{3}{c}{Tracking(ms)} & \multicolumn{3}{c}{Pose estimate(ms)} & \multicolumn{3}{c}{Total(ms)}\\
			& VINS-Mono & OpenVINS & SP-VIO & VINS-Mono & OpenVINS & SP-VIO & VINS-Mono & OpenVINS & SP-VIO \\
			\midrule
			MH\_01\_easy  & 7.31 & 3.02  & 3.02 & 29.92  & 3.54 & 4.87 & 36.23 & 6.56 & 7.89\\
			MH\_02\_easy  & 7 & 3.02 & 2.75 & 30.13 & 3.63 & 4.49 & 37.13 & 6.65 & 7.24\\
			MH\_03\_medium  & 6.9 & 2.86 & 2.82 & 29.24 & 3.73 & 4.56 & 36.14 & 6.59 & 7.38\\
			MH\_04\_difficult  & 6.81 & 2.93 & 2.74 & 28.31 & 3.57 & 4.26 & 35.13 & 6.5 & 7\\
			MH\_05\_difficult  & 6.73 & 3.05 & 2.92 & 29.12 & 3.7 & 4.42 & 35.85 & 6.75 & 7.34\\
			V1\_01\_easy  & 6.84 & 2.62 & 2.74 & 30.75 & 3.91 & 4.54 & 37.59 & 6.53 & 7.28\\
			V1\_02\_medium  & 7.16 & 2.86 & 2.75 & 20.08 & 3.67 & 3.86 & 27.24 & 6.53 & 6.61\\
			V1\_03\_difficult  & 7.71 & 3.12 & 3.12 & 19.83 & 3.39 & 3.58 & 27.54 & 6.51 & 6.7\\
			V2\_01\_easy  & 6.76 & 2.59 & 2.68 & 29.41 & 3.36 & 4.09 & 36.17 & 5.95 & 6.77\\
			V2\_02\_medium  & 7.12 & 3.03 & 2.75 & 22.91 & 3.53 & 3.75 & 30.03 & 6.56 & 6.5\\
			V2\_03\_difficult  & 10.13 & 3.64 & 3.55 & 15.53 & 2.74 & 2.83 & 25.66 & 6.38 & 6.38\\
			Average & 7.31  & 2.97  & 2.89  & 25.93  & 3.52  & 4.11  & 32.24  & 6.49  & 7 \\
                \midrule
                Room1\_512\_16  & 11.78  & 3.04  & 2.97  & 16.33  & 3.03  & 3.19  & 28.11  & 6.07  & 6.16\\
			Room2\_512\_16  & 11.33  & 2.74  & 2.68  & 18.12  & 3.02  & 2.99  & 29.45  & 5.76  & 5.67\\
			Room3\_512\_16  & 12.08  & 2.95  & 2.85  & 16.03  & 2.98  & 3.00  & 28.11  & 5.93  & 5.85\\
			Room4\_512\_16  & 11.31  & 2.97  & 2.97  & 16.77  & 3.19  & 3.36  & 28.08  & 6.16  & 6.33\\
			Room5\_512\_16  & 11.87  & 3.23  & 3.27  & 14.40  & 2.67  & 3.23  & 26.27  & 5.90  & 6.50\\
			Room6\_512\_16  & 10.89  & 2.40  & 2.48  & 21.17  & 2.95  & 3.54  & 32.06  & 5.35  & 6.02\\
			Average & 11.54  & 2.88  & 2.87  & 17.14  & 2.97  & 3.21  & 28.68  & 5.85  & 6.08\\
			\bottomrule
		\end{tabular}
	}

\end{table*}

\subsection{Performance Evaluation under Discontinuous Observation Conditions}\label{sec:Performance evaluation under discontinuous observation conditions}
In this subsection, we choose the UAV flight data in Nudt VI to verify the robustness of SP-VIO. 
This scene is a large-scale flight experiment in outdoor environment, which does not have the condition of loop correction and can only realize the performance of VIO.
As described in \secref{sec:Mainstream VIO Algorithms}, the correction strategies of VINS-Mono and ORB-SLAM3 are consistent in the scenario without visual loop closure, so only VINS-Mono is used as the comparison algorithm.
Notably, OpenVINS and MSCKF have the same processing strategy for visual deprived conditions, so it is not used as a comparison algorithm.

\begin{figure}[htbp]
	\centering
	\includegraphics [width=3.5in]{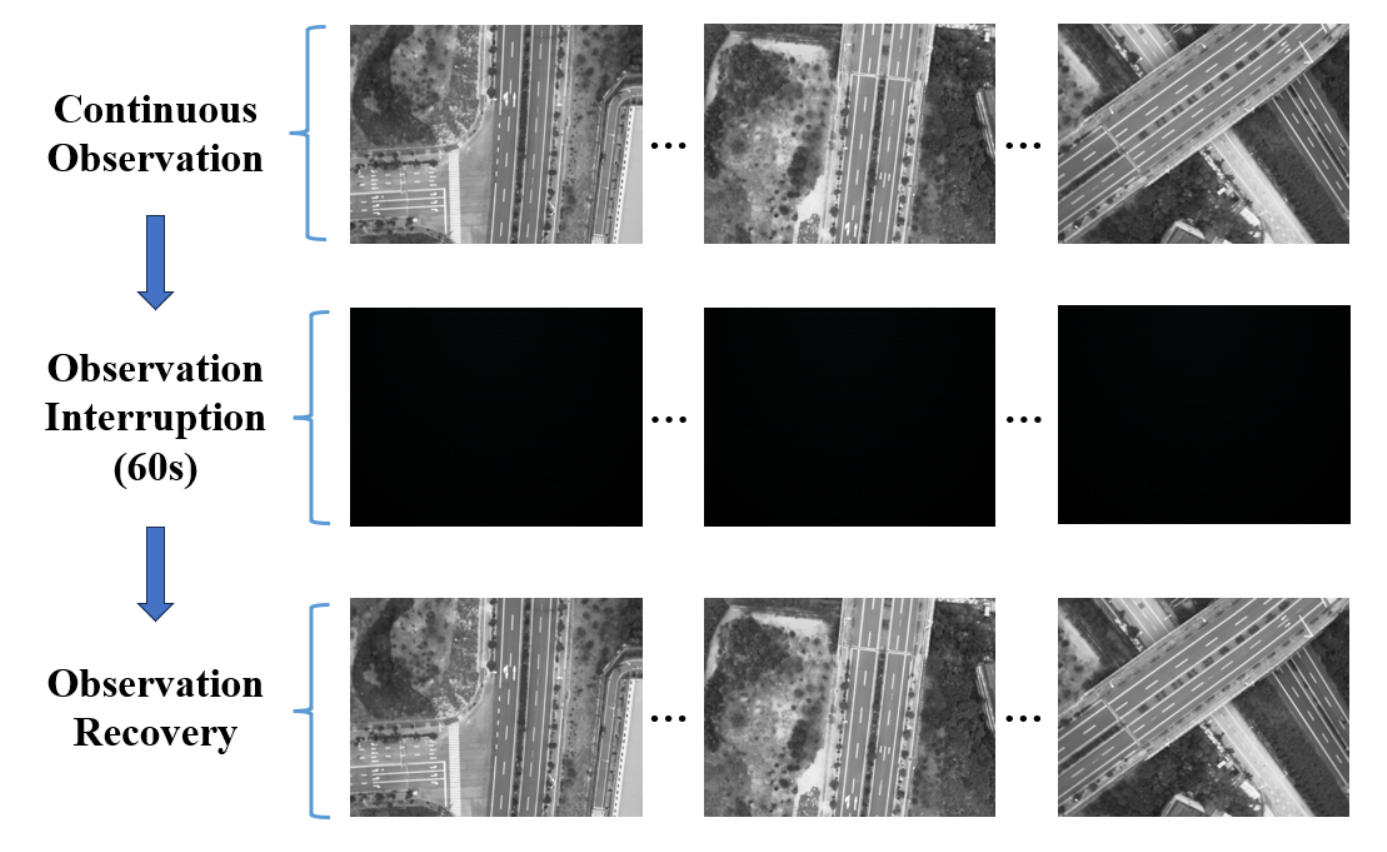}\label{fig:interruption diagram}  
	\caption{
		Visual observation interruption diagram (continuous observation, observation interruption, observation recovery).
	}
	\label{fig:Visual observation interruption diagram T}
\end{figure}

The specific setting of visual deprivation is that the input image is replaced with occluded image at random time, which lasts for 60s and then is restored. The schematic diagram of the process is shown in \figref{fig:Visual observation interruption diagram T}.

\begin{figure*}[htbp]
	\centering
	\subfigure[] { \centering \includegraphics [width=3.5in] {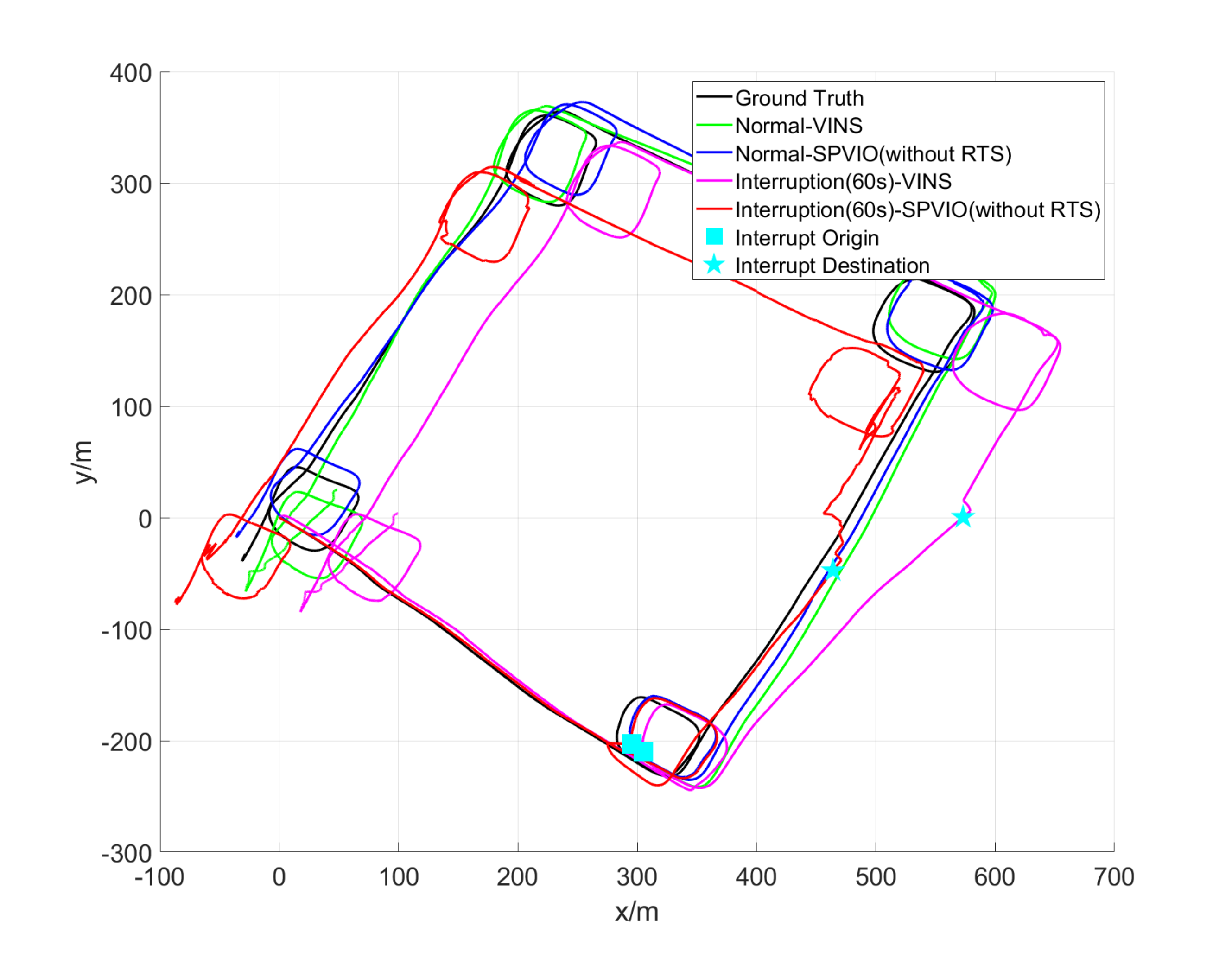}\label{fig:Without RTS backtracking correction}} 
	\subfigure[] { \centering \includegraphics [width=3.5in] {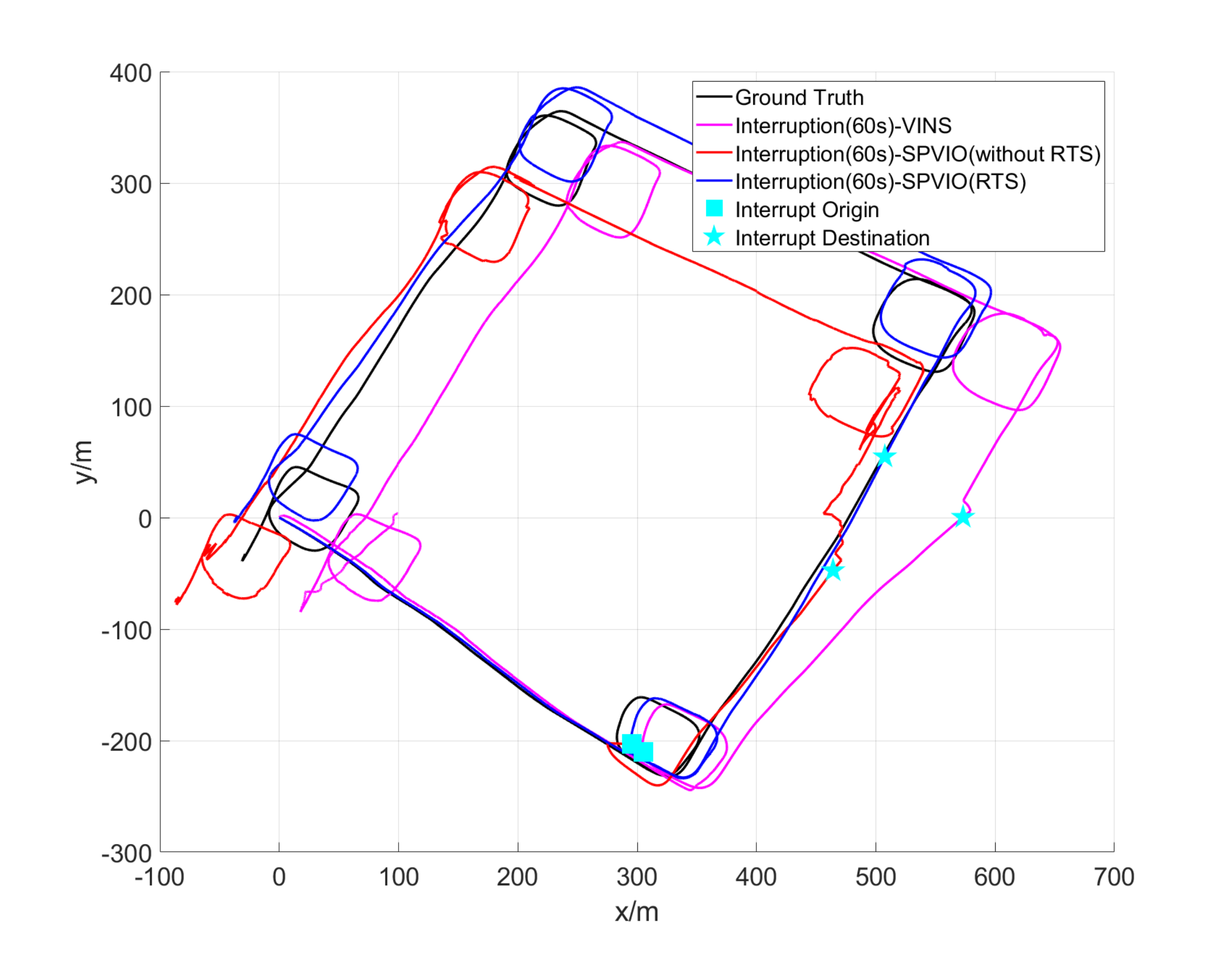}\label{fig:With RTS backtracking correction}}
	\caption{
		Comparison of changes in the result trajectories after partial interruption of visual observation
		(a) Without RTS backtracking correction.
		(b) With RTS backtracking correction.
	}
	\label{fig:Partial result trajectories in public datasets1}
\end{figure*}

As shown in \figref{fig:Without RTS backtracking correction}, although the state estimation of the two algorithms reconverges after observation recovery, the cumulative error cannot be eliminated due to the lack of the function of backtracking correction, which affects the accuracy of subsequent navigation results.
After RTS backtracking correction, the localization error trend of SP-VIO during visual interruption is slowed down, so that the system can obtain a better navigation result after visual information recovery as shown in \figref{fig:With RTS backtracking correction}.

To show the navigation results under different observation conditions more intuitively, we summarize the absolute pose errors in \tabref{tab:Discontinuous Observation Experiment}. 
After adopting the RTS repositioning algorithm, SP-VIO's error growth percentage under discontinuous observation conditions is reduced to $35.08\%$, which is better than that of VINS-Mono ($77.56\%$).

\begin{table}[htbp]
	\begin{center}
		\setlength{\tabcolsep}{5pt}
		\caption{RMSE in Discontinuous Observation Experiment in Meters}
		\label{tab:Discontinuous Observation Experiment}
	\begin{threeparttable}	
		\begin{tabular}{*{4}{c}}
			\toprule
			Algorithm & Continuous & Lost\_60s & Percentage$^{1}$ \\			
			\midrule
			VINS-Mono & $27.77$ & $49.31$ & $77.56\%$ \\
			SP-VIO(without RTS) & $\boldsymbol{17.22}$ & $59.69$ & $246.63\%$ \\
			SP-VIO(RTS) & $\boldsymbol{17.22}$ & $\boldsymbol{23.26}$ & $35.08\%$ \\
			\bottomrule
		\end{tabular}
		\begin{tablenotes}
			\footnotesize
			\item[1] This percentage is about localization error growth after 60 seconds of observation interruption.
		\end{tablenotes}
	\end{threeparttable}	
	\end{center}
\end{table}

In general, SP-VIO has excellent positioning performance and robustness, especially in challenging discontinuous observation environment can also obtain high-precision navigation results.

\section{Conclusion and Discussion}
This paper proposes the SP-VIO that takes into account accuracy, efficiency and robustness. The specific advantages are as follows:  
(1) It maintains the efficiency advantage of filter-based VIO;
(2) Through the reconstruction of the system model and measurement model, the accuracy and consistency of our algorithm are improved;
(3) The accumulated errors during visual deprivation can be corrected under non-closed loop conditions, with better robustness.
The experiments based on public and personal datasets demonstrate that SP-VIO has better localization accuracy and robustness than SOTA VIO algorithms while remaining efficient.

In future research work, we will further explore the extension of our algorithm in SLAM and the deployment application on payload-constrained robots to obtain a more comprehensive and powerful visual-inertial navigation system.

\bibliographystyle{IEEEtran}

\bibliography{Ref}

\vfill

\end{document}